\documentclass[11pt]{article}
\usepackage{amsmath}

\usepackage[final]{acl}

\usepackage{times}
\usepackage{latexsym}

\usepackage[T1]{fontenc}
\usepackage[utf8]{inputenc}

\usepackage{microtype}

\usepackage{inconsolata}

\usepackage{graphicx}

\usepackage{multirow, amssymb}
\usepackage{booktabs}
\usepackage{makecell}
\usepackage{pifont}
\usepackage{xcolor}
\usepackage{url}
\usepackage{algorithm}
\usepackage{algorithmicx}
\usepackage{algpseudocode}
\floatname{algorithm}{Algorithm} %算法

\definecolor{CheckGreen}{HTML}{2E7D32}
\definecolor{CrossRed}{HTML}{C62828}
\definecolor{PartAmber}{HTML}{F39C12}

\newcommand{\chkmark}{\textcolor{CheckGreen}{\ding{51}}}  % ✓
\newcommand{\crossmark}{\textcolor{CrossRed}{\ding{55}}}  % ✗
\newcommand{\ours}{\texttt{PaperScope}}

\title{PaperScope: A Multi-Modal Multi-Document Benchmark for Agentic Deep Research Across Massive Scientific Papers}

\author{
Lei Xiong$^{1,2}$,
Huaying Yuan$^{1}$,
Zheng Liu$^{2}$\thanks{Corresponding author},
Zhao Cao$^{1}$,
Zhicheng Dou$^{1}$\footnotemark[1]\\
$^{1}$Gaoling School of Artificial Intelligence, Renmin University of China, Beijing, China\\
$^{2}$Beijing Academy of Artificial Intelligence, Beijing, China\\
{\tt\small xiongxiongleilei@ruc.edu.cn}
}

\begin{document}
\maketitle
\begin{abstract}
Leveraging Multi-modal Large Language Models (MLLMs) to accelerate frontier scientific research is promising, yet how to rigorously evaluate such systems remains unclear. Existing benchmarks mainly focus on single-document understanding, whereas real scientific workflows require integrating evidence from multiple papers, including their text, tables, and figures. As a result, multi-modal, multi-document scientific reasoning remains underexplored and lacks systematic evaluation.
To address this gap, we introduce \ours{}, a multi-modal multi-document benchmark designed for agentic deep research. \ours{} presents three advantages:
(1) \textbf{Structured scientific grounding.}
It is built on a knowledge graph of over 2,000 AI papers spanning three years, providing a structured foundation for research-oriented queries.
(2) \textbf{Semantically dense evidence construction.}
It integrates semantically related key information nodes and employs an optimized random-walk article selector to sample thematically coherent paper sets, thereby ensuring adequate semantic density and task complexity.
(3) \textbf{Multi-task evaluation of scientific reasoning.}
It contains over 2,000 QA pairs across reasoning, retrieval, summarization, and problem solving, enabling evaluation of multi-step scientific reasoning.
Experimental results show that even advanced systems such as OpenAI Deep Research and Tongyi Deep Research achieve limited scores on \ours{}, highlighting the difficulty of long-context retrieval and deep multi-source reasoning. \ours{} thus provides a rigorous benchmark alongside a scalable pipeline for constructing large-scale multi-modal, multi-source deep research datasets. The code and dataset are available at: https://github.com/CherYou/PaperScope.

\end{abstract}

\section{Introduction}
\label{sec:intro}

Scientific papers are inherently multi-modal, and solving complex scientific problems demands retrieving, reasoning, and synthesizing information across multiple documents that span text, tables, figures, formulas, and algorithms. Agentic deep research systems~\cite{dr,google_dr,tongyidr} have reshaped research workflows by autonomously planning retrieval strategies, invoking external tools, refining queries adaptively, and verifying context, thereby exhibiting stronger dynamic reasoning~\cite{geng2025webwatcher,Li2025webthinker,jin2025search}than traditional RAG for research‑style tasks. 

However, current systems remain primarily grounded in internet text resources and often overlook large volumes of stored multi-modal documents such as papers, technical reports, financial reports, and lab manuals that encode dense, structured visual and symbolic information~\cite{dong2025doc}; failing to leverage these artifacts limits the applicability and knowledge coverage in realistic scientific pipelines. Existing benchmarks also fall short: document understanding datasets mainly evaluate single-page or multi-page single‑document  tasks~\cite{chartqa,docvqa,cui2025curie,infovqa,li2024mmsci,mmlongbench,arxivqa,li2024m3sciqa,tian2025mmcr,li2025deepsolution,wang2024charxiv}, and agent benchmarks largely target general‑purpose assistants and web browsing~\cite{bc_en,mialon2023gaia,hle}. As a result, multi-modal, multi‑document scientific reasoning, especially in large‑file environments, remains underrepresented, and there is a pressing need for realistic, reproducible protocols that faithfully reflect research workflows.

To address these gaps, we introduce \ours{}, a multi-modal, multi‑document agentic benchmark for scientific deep research. \ours{} introduces the following key features: 

(1) \textbf{Assess agents’ capabilities in multi-source information retrieval, synthesis, reasoning, and generation.}
    Each question is associated with a large corpus, where answers are dispersed across heterogeneous sources and modalities. The benchmark corpus contains 202–1500 documents spanning multiple sub-fields of artificial intelligence, including reinforcement learning, generative models, and computer vision---thus covering a broad range of scientific research scenarios in AI.

(2) \textbf{Provide diverse task formats across scientific deep-research skills.}
    It includes Reasoning, Topic Induction, Summary, and Solution tasks, collectively capturing document retrieval, document understanding, multi-source information integration, and methodological formulation. The questions include both objective and subjective types; this diversity increases the evaluation difficulty and more faithfully reflects the complexity of real-world research applications.

(3) \textbf{Ensure high-quality, graph-grounded annotations.}
    Following an inverted-construction strategy, we build each question using document sets sampled through random walks on a knowledge graph constructed from top AI conferences, ensuring answer accuracy and near-uniqueness within the corpus-level knowledge graph. We further perform strict quality control on both documents and annotations to guarantee task specificity and correctness across all categories.

We conduct comprehensive experiments on \ours{}, evaluating two categories of models: standard ReAct-based agents and Deep Research agents for a total of 16 systems. Even leading open-source and closed-source agents fail to achieve strong performance. Our benchmark reveals that existing models still face substantial limitations in large-scale document deep-research scenarios: their accuracy remains insufficient, and their multi-modal understanding, retrieval, and multi-source information integration capabilities require significant improvement.

In summary, our contributions are threefold:

(1) We introduce \ours{}, a multi-modal, multi-document benchmark for scientific deep research, targeting large-scale retrieval, cross-document reasoning, and multi-source information synthesis under realistic research settings.

(2) \ours{} is built from a heterogeneous corpus via a knowledge-graph-guided construction pipeline, providing fine-grained, near-unique annotations and a reproducible methodology for creating multi-source, multi-modal, multi-document deep-research datasets.

(3) Experiments on 16 state-of-the-art ReAct-based and Deep Research agents reveal substantial performance gaps in accuracy, multi-modal understanding, retrieval, and information integration, highlighting the difficulty of large-scale scientific deep research.

\section{Related Work}
\begin{table*}[t]
\centering
\small
\setlength{\tabcolsep}{4pt}
\resizebox{\textwidth}{!}{%
\begin{tabular}{l|c|c|c|c|c|c|c|c}
\toprule
\textbf{Benchmarks} & \textbf{Avg. Papers} & \textbf{Avg. Pages} & \textbf{Cross-Page} & \textbf{Multi-Modal} & \textbf{Multi-Doc} & \textbf{Capability} & \textbf{Open Form} & \textbf{Agent} \\
\midrule
\multicolumn{9}{c}{\textbf{LSDU Benchmark}} \\
\midrule
DocVQA & - & 1 & \crossmark & \crossmark & \crossmark & U & \crossmark & \crossmark \\
ChartQA & - & 1 & \crossmark & \crossmark & \crossmark & U & \crossmark & \crossmark \\
InfoVQA & - & 1 & \crossmark & \crossmark & \crossmark & U & \crossmark & \crossmark \\
Charxiv & - & 1 & \crossmark & \crossmark & \crossmark & U & \crossmark & \crossmark \\
ArxivQA & - & 1 & \crossmark & \crossmark & \crossmark & U & \crossmark & \crossmark \\
MMSCI & - & 1 & \crossmark & \crossmark & \crossmark & U & \crossmark & \crossmark \\
MMLongBench-Doc & 1 & 47.5 & \chkmark & \crossmark & \crossmark & U, SRea & \crossmark & \crossmark \\
MMCR & 1 & 19 & \chkmark & \chkmark & \crossmark & U, MRea & \crossmark & \crossmark \\
CURIE & 1 & - & \chkmark & \chkmark & \crossmark & U & \crossmark & \crossmark \\
M3SciQA & 2 & - & \chkmark & \chkmark & \chkmark & Retr, U & \crossmark & \crossmark \\
DeepSolution & 1 & - & \chkmark & \crossmark & \crossmark & U & \chkmark & \crossmark \\
\midrule
\multicolumn{9}{c}{\textbf{Agentic Deep Research Benchmark}} \\
\midrule
HLE & - & - & - & \crossmark & \crossmark & Rea, Think, G & \crossmark & \chkmark \\
BrowseComp & - & - & - & \crossmark & Web & Retr, MRea, Think & \crossmark & \chkmark \\
GAIA & - & - & - & \crossmark & Web & Retr, MRea, Solu, Think, G & \chkmark & \chkmark \\
M4DocBench & 3.8 & 7 & \chkmark & \chkmark & \chkmark & Retr, U, MRea, Think, G & \chkmark & \chkmark \\
\textbf{PaperScope Bench} & \textbf{$>=$500} & \textbf{$>$5000} & \textbf{\chkmark} & \textbf{\chkmark} & \textbf{\chkmark} & \textbf{Retr, U, MRea, Solu, Think, G} & \textbf{\chkmark} & \textbf{\chkmark} \\
\bottomrule
\end{tabular}%
}
\caption{
Comparison of existing benchmarks and our proposed \textbf{Agentic Deep Research Benchmark}. 
Symbols: \chkmark~indicates support, \crossmark~indicates lack of support. 
Abbreviations: U = Understanding, SRea = Single-resource reasoning, MRea = Multi-resource reasoning, Retr = Retrieval, Solu = Solution, Think = Thinking, G = Generation.
}
\label{tab:benchmark_comparison}
\end{table*}
\paragraph{Long Scientific Document Understanding Benchmarks.} 
Existing benchmarks for long scientific document understanding primarily focus on single-document settings and remain limited in cross-page or multi-source reasoning. Charxiv~\cite{wang2024charxiv} and ArxivQA~\cite{arxivqa} construct figure-centric QA with limited grounding in full-paper content, while MMSCI~\cite{li2024mmsci} is restricted to caption-based tasks. CURIE~\cite{cui2025curie} extends to scientific problem solving across multiple domains but still operates within individual documents. MMCR~\cite{tian2025mmcr} and M3SciQA~\cite{li2024m3sciqa} introduce cross-source or citation-based settings, yet do not support genuine multi-document synthesis. DeepSolution~\cite{li2025deepsolution} evaluates solution generation from structured PDF content, and M4DocBench~\cite{dong2025doc} targets multi-modal and multi-turn reasoning under limited human annotations, but neither emphasizes multi-document retrieval or complex problem solving.
Overall, existing datasets remain largely confined to single papers or single sources, leaving cross-page, multi-modal, and multi-document scientific retrieval, reasoning, summarization, and problem solving insufficiently explored.

\paragraph{General Agentic Deep Research Benchmarks.}
Agentic deep research systems extend traditional RAG through autonomous retrieval planning, adaptive tool use, and contextual verification. However, existing agent benchmarks—such as HLE, BrowseComp, and GAIA~\cite{hle,bc_en,gaia}—primarily target web-based general-purpose information seeking and offer limited coverage of scientific documents rich in symbolic, visual, and algorithmic structures. Their tasks rarely require integrating multiple local files or synthesizing multi-modal evidence, leaving key aspects of real scientific workflows untested. \ours{} Bench addresses these gaps by constructing high-correlation scientific subsets and emphasizing cross-document retrieval, multi-modal grounding, and multi-resource reasoning, enabling a unified evaluation of retrieval, understanding, reasoning, and solution generation.

\section{PaperScope Bench}
\begin{figure*}[h]
    \centering
     %\vspace{-2ex}
    \includegraphics[width=1\linewidth]{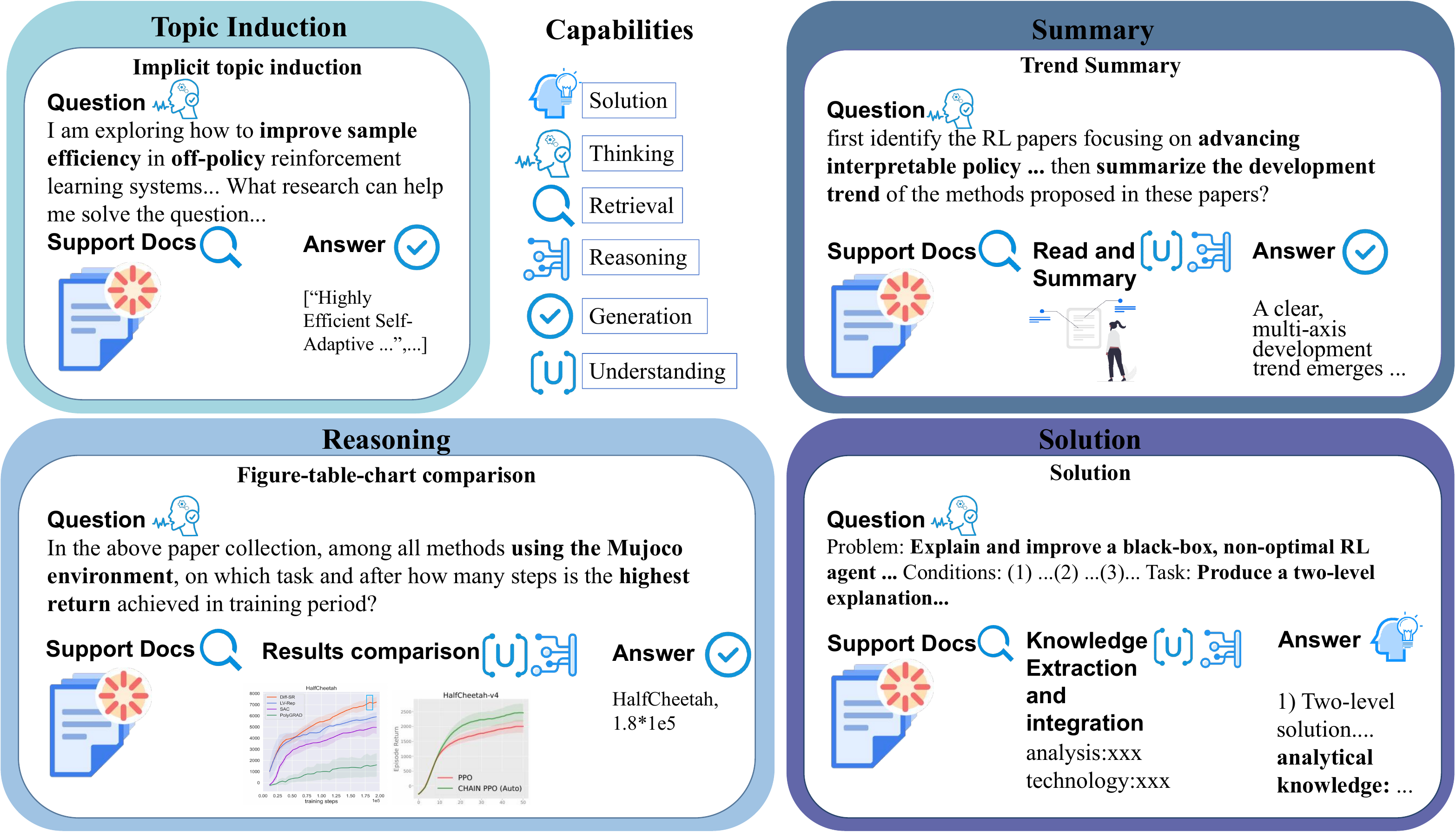}
    \caption{Visualized Examples of \ours{} Bench: Sub-task illustrations from four meta-tasks. The icons in the center represent the various capabilities required by the agent. In each case, the icons placed next to specific stages indicate the particular capabilities needed at that stage. Thinking refers to the reasoning and decomposition of the underlying intent of a given query. Understanding denotes multi-modal comprehension of the document content.}
    %\vspace{-2ex}
    \label{fig:datasetcase}
\end{figure*}
\subsection{Overview}
We propose \ours{}, a multi-modal benchmark designed to evaluate Deep Research agents across a large-scale scientific corpus. As outlined in Table~\ref{tab:task_stat} and Figure~\ref{fig:datasetcase}, the benchmark comprises 2,400 questions spanning 11 sub-tasks. These are organized into four meta-categories—\textit{Topic Induction}, \textit{Multi-Document Reasoning}, \textit{Summary}, and \textit{Solution}—which are structured to mirror the hierarchical research workflow from retrieval to synthesis. Detailed task statistics are provided in Appendix~\ref{sec:appendix overview}.
\subsection{Data Collection}
To evaluate retrieval, scientific understanding, synthesis, and problem solving across a large-scale scientific corpus, we construct a multi-modal, multi-document deep-research dataset built from scientific papers. Existing benchmarks rarely test complex multi-document retrieval and reasoning, motivating our new design. We collect 25,495 papers from ArXiv\footnote{arXiv: https://arxiv.org}
 and OpenReview\footnote{openreview: https://openreview.net}
, covering top AI conferences from 2023–2025 across more than 20 venues. All PDFs are rigorously filtered for completeness, readability, and quality, ensuring high data integrity and domain relevance. This curation allows \ours{} Bench to capture the scale, diversity, and complexity of real scientific research scenarios.

\subsection{Task Creation}
\paragraph{Topic Induction Tasks.}
This task evaluates a model’s ability to perform multi-modal semantic retrieval and topic recognition within a large scientific corpus. Given a textual or multi-modal query, the model must retrieve relevant papers under two settings: (1) \emph{Implicit Topic Induction}, where the query provides latent semantic cues and the model must infer the underlying theme; (2) \emph{Explicit Topic Induction}, where the topic is directly specified and the model must identify the most relevant studies.

\paragraph{Reasoning Tasks.}
This task measures cross-document reasoning and multi-modal evidence integration---capabilities central to scientific deep research. After retrieving related papers, the model must integrate textual and visual information across sources. Five sub-tasks are included: (1) \emph{Figure–Table–Chart Comparison}, comparing visual and quantitative findings across papers; (2) \emph{Figure–Table–Chart Reasoning}, integrating heterogeneous visual evidence for logical inference; (3) \emph{Formula Reasoning}, interpreting mathematical expressions and variable relationships; (4) \emph{Algorithm Reasoning}, understanding pseudocode and procedural logic; (5) \emph{Full-Paper Reasoning}, synthesizing semantic and structural information across full texts.

\paragraph{Summary Tasks.}
This task assesses the ability to integrate knowledge across multiple papers and produce higher-level scientific abstractions. Given a thematic area and related documents, the model must generate structured, coherent summaries across three sub-tasks: (1) \emph{Trend Summary}, characterizing research trajectories and emerging directions; (2) \emph{Method Summary}, summarizing strengths and weaknesses of related methods; (3) \emph{Fine-Grained Summary}, comparing experimental results or metrics across studies.

\paragraph{Solution Tasks.}
Inspired by DeepSolution~\cite{li2025deepsolution}, this category evaluates scientific problem solving and innovation. Beyond retrieval and understanding, the model must integrate multi-modal and multi-document evidence to propose actionable, well-grounded solutions. The single sub-task, \emph{Solution Generation}, requires retrieving relevant literature, extracting cross-modal evidence, and producing a comprehensive solution with explicit use of figures, algorithms, and results.
\subsection{Statistics of Benchmark}
\ours{} comprises a total of 2,400 multi-modal, multi-document questions distributed across four major categories. Specifically, the number of questions for each task and sub-task is summarized in Table~\ref{tab:task_stat}. Further details on dataset composition, sampling, and annotation procedures are provided in Appendix~\ref{sec:appendix overview} and~\ref{sec:appendix dataconstruction}.
\begin{table}[t]
\centering
\small
\setlength{\tabcolsep}{2pt}
\renewcommand{\arraystretch}{0.95}
\resizebox{\columnwidth}{!}{%
\begin{tabular}{llc}
\toprule
\textbf{Category} & \textbf{Sub-Task} & \textbf{Num} \\
\midrule
\multirow{5}{*}{Multi-Doc Reasoning Task} 
& Figure-table-chart comparison & 200 \\
& Figure-table-chart reasoning & 200 \\
& Formula reasoning & 200 \\
& Algorithm reasoning & 200 \\
& Full paper reasoning & 200 \\
\midrule
\multirow{2}{*}{Topic Induction Task} 
& Implicit topic induction & 200 \\
& Explicit topic induction & 200 \\
\midrule
\multirow{3}{*}{Summary Task} 
& Trend summary & 200 \\
& Method summary & 200 \\
& Fine-grained summary & 200 \\
\midrule
Solution Task & Solution generation & 400 \\
\bottomrule
\end{tabular}
}%
\caption{Statistics of task categories and subtasks.}
\label{tab:task_stat}
\end{table}
%\vspace{-5 mm}  % 缩小表格底部空白

\section{Bench Construction Methodology}
\subsection{Overview}

\begin{figure}[t]
    \centering
    \includegraphics[width=0.47\textwidth]{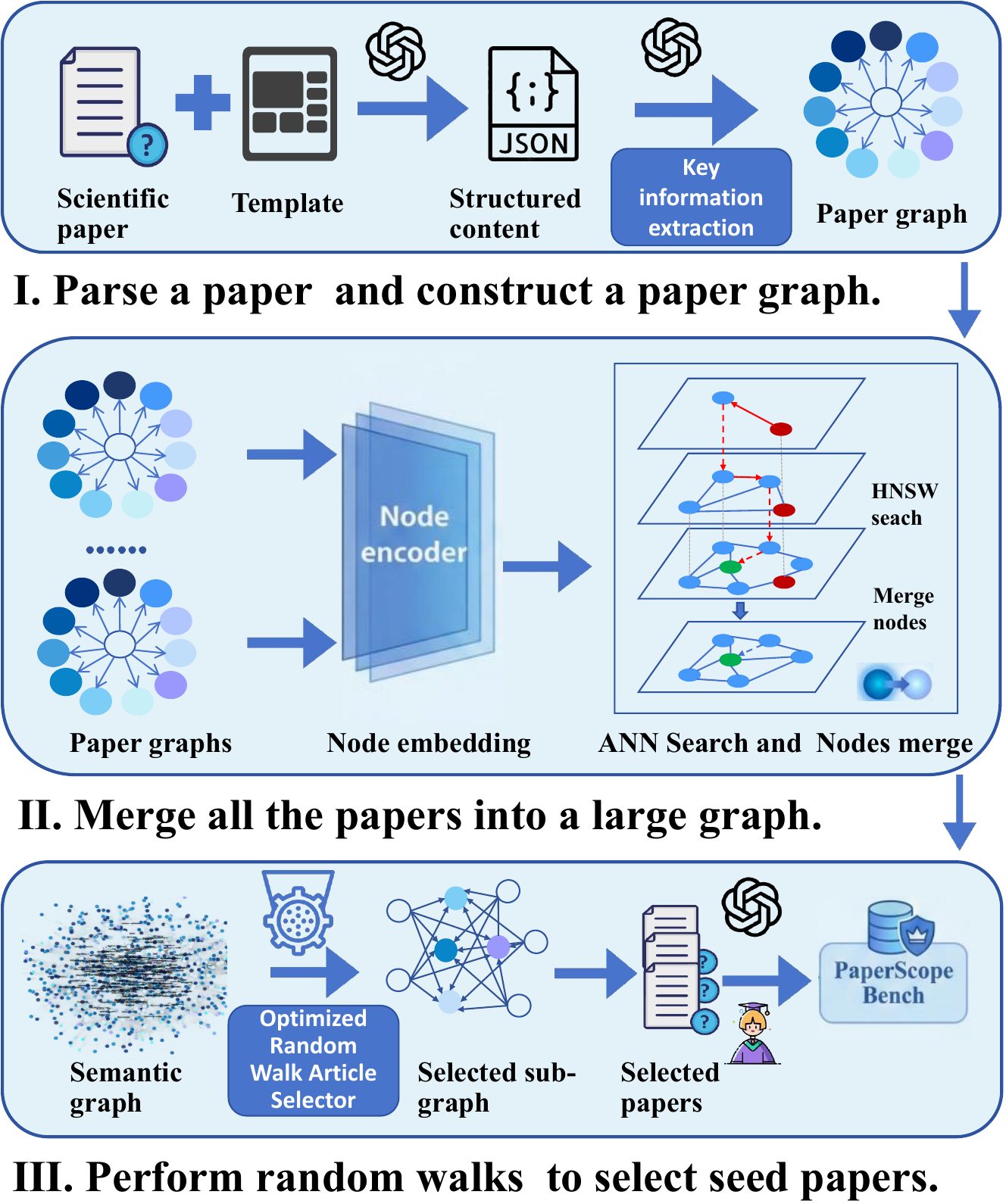}
    %\vspace{-1ex}
    \caption{The overview of construction methodology of the \ours{} Bench.}
    %\vspace{-2ex}
    \label{fig:benchconstruction}
\end{figure}

In order to ensure the relevance of the seed papers collection and the difficulty level of the constructed problems, we select the seed paper collection by constructing a knowledge graph of the local large documents and employing random walks. We propose a heterogeneous large-graph-based multi-document association framework for scientific paper selection, designed to ensure answer uniqueness and fully leverage multi-modal evidence in multi-source scenarios. The framework consists of three stages, which is illustrated in Figure~\ref{fig:benchconstruction}: (1) document-level key information extraction and graph construction; (2) efficient nearest-neighbor retrieval and semantic disambiguation merging based on the Hierarchical Navigable Small World (HNSW) graph~\cite{malkov2018efficient}, which enables logarithmic-time approximate nearest-neighbor search through a multi-layer small-world topology; (3) performing Optimized Random Walk-based Article Selection (ORWAS) on large graphs to identify high-quality papers sharing common key information nodes, which are then used to construct \ours{}.
\subsection{Graph Construction}
For every paper, we extract structured key information nodes with a large language model, following prior evidence that LLMs perform well on scientific information extraction~\cite{dagdelen2024structured}. We consider 13 key information nodes: \textit{title, research background, classification tags, key contributions, methodology, datasets, results, metrics, formulas, algorithms, figures, tables, and limitations}. Each paper becomes a title node connected to its key information nodes via ``belongs‑to'' edges.
\subsection{Semantic Disambiguation and Merging}

To ensure the relevance of the documentation, we merge the knowledge graphs of each paper. Key information nodes across papers are linked when they are semantically related. All nodes are embedded into a 4096‑dimensional space using a shared encoder with type embeddings to accommodate heterogeneous content, including text, figures, tables, algorithms, and formulas. To consolidate key information nodes globally and efficiently, we index all node embeddings with HNSW~\cite{malkov2018efficient} and retrieve Top‑K neighbors per node. Similarities are computed only within the same key information type, and nodes exceeding a threshold $\theta$ are merged while preserving provenance for traceability. We adopt a coarse‑to‑fine schedule: frequent and clearer types (e.g., classification tags and datasets) are merged first, followed by semantically richer types (e.g., figures, algorithms, and formulas). Typical HNSW settings use M=32, efConstruction=50, and efSearch=30 with K=20.
\subsection{Optimized Random-Walk Article Selector (ORWAS)}
The selection process operates on a heterogeneous graph (approx. 30,000 nodes, 200,000 edges). To ensure efficiency and quality, we employ the following strategies:
(1) \textbf{Graph Optimization:} Adjacency pre-indexing and neighbor caching reduce access latency to near-constant time.
(2) \textbf{Sampling \& Walk Strategy:} Initialization is stratified (70\% article nodes, 30\% high-frequency key information nodes). We use a structure-aware random walk: transitions from paper nodes are biased towards high-frequency key information nodes, while others are uniform.
(3) \textbf{Scalability:} We implement batched parallelization, local accumulation, and strict memory caps on candidate size to prevent overload.
(4) \textbf{Ranking:} Candidates undergo constrained enumeration and are ranked via a composite score (coverage, diversity, consistency, and redundancy). Details are in Appendix~\ref{sec:appendix dataconstruction}.

Based on the paper sets and shared key information nodes identified by ORWAS, we construct the benchmark using 4 distinct strategies (details in Appendix~\ref{sec:appendix dataconstruction}).

\subsection{Quality Control}
\label{quality control}
\textbf{Answer Uniqueness.} Adopting BrowseComp’s~\cite{bc_en} inverted formulation strategy, we derive questions from known facts to converge on unique correct answers, thereby minimizing open-world ambiguity and ensuring verifiability.

\textbf{Multi-modal Dependence.} To prevent textual shortcuts, key evidentiary cues are exclusively embedded in visual modalities (figures, tables, layouts). 

\textbf{Intrinsic Difficulty.} We enforce difficulty via a robustness screen: questions are retained only if SOTA models (Gemini 2.5-Pro~\cite{gemini_25_pro}, GPT-5~\cite{gpt5}) fail to solve them given a single web-search attempt.

\section{Experiments}
\subsection{Evaluation}
\ours{} includes 4 task categories that evaluate the end-to-end scientific deep research agent from retrieval and understanding to synthesis and problem solving. Each task uses tailored metrics, with detailed formulas and prompts provided in Appendix~\ref{sec:appendix eval}.

(1) Topic Induction: We report \textbf{Recall@K} to measure coverage and topic localization. (2) Reasoning: Performance is measured using \textbf{Exact Match (EM)}, reflecting the reliability of multi-source reasoning. (3) Summary: We use a hybrid evaluation combining induction scores with a GPT-5 score across five dimensions: \textbf{fluency, relevance, accuracy, creativity, and overall quality}, following ResearchPulse~\cite{chen2025researchpulse}. (4) Solution: Following SolutionBench~\cite{li2025deepsolution}, we use \textbf{Analysis and Technical Scores} to assess problem decomposition, reasoning design, and domain-knowledge usage, with scores judged by GPT-5~\cite{gpt5}.

\begin{table*}[!htbp]
\centering
\small
\renewcommand{\arraystretch}{1.2} % 更紧凑的行间距
\begin{tabular}{l|l|r r r r r}
\toprule
\textbf{Agent Type} & \textbf{Model} & \textbf{Reasoning} & \textbf{Induction} & \textbf{Summary} & \textbf{Solution} & \textbf{Score} \\
\midrule
\multirow{11}{*}{LLM-based ReAct} 
& WebWatcher 32B& 4 & 0 & 46.74 & 26.78 & 18.70 \\
& OpenAI 4o-mini & 6 & 25.49 & 53.26 & 22.1 & 23.74 \\
& Gemini-2.5-flash-thinking & 7 & 13.33 & 38.40 & 29.71 & 19.32 \\
& OpenAI GPT-5.1 & 0 & 0 & 42 & 51.84 & 17.78 \\
& Gemini 2.5 pro & 3 & 7.02 & 47.54 & 40.39 & 20.50 \\
& GLM 4.5V & 0 & 0 & 37.32 & 32.45 & 14.44 \\
& Kimi k2 & 12 & 24.07 & \textbf{56.64} & 49.85 & 30.38 \\
& Qwen3-VL & 4 & 13.33 & 52.74 & 37.38 & 22.89 \\
& deepseek-V3.1 & 6 & \textbf{26.32} & 52.22 & 51.65 & 26.46 \\
\midrule
\multirow{7}{*}{Deep Research} 
& DR Tulu-8B & 4 & 0 & 40.60 & 38.71 & 18.05 \\
& MMSearch-R1-7B & 8 & 3.70 & 43.66 & 17.21 & 19.19 \\
& ASearcher-Web-7B & 13 & 0 & 47.26 & 8.95 &  21.57 \\
& MiroThinker-v1.0-30B & 3 & 3.92 & 27.44 & 32.03 & 13.33 \\
& Tongyi Deep Research 30B & 11 & 0 & 5 & 36.55 & 10.66 \\
& OpenAI o3 deep research& 13 & 0 & 56.26 & \textbf{59.15} & 29.29 \\
& Grok 4 & \textbf{36} & 20 & 53.74 & 48.28 & \textbf{40.95} \\

\bottomrule
\end{tabular}
\caption{Performance of different agents across Reasoning, Induction, Summary, and Solution tasks. The best score on each dataset is shown in bold.}
\label{tab:agent_performance}
\end{table*}

\subsection{Settings and Baselines}
\subsubsection{Experiment Settings}
We evaluate agents in a unified multi-modal retrieval and reasoning environment for multi-document scientific tasks. Inputs include high-resolution images and structurally parsed PDF–Markdown documents (processed with MinerU2.5~\cite{niu2025mineru2}), enabling joint text–image reasoning over figures, formulas, and pseudocode.

Models follow the ReAct~\cite{yao2022react} paradigm, alternating between reasoning and action. They plan via chain-of-thought, invoke tools for retrieval and parsing, and generate final answers. All models use two tools: (i) \textbf{Local FileSearch} for semantic retrieval using an Ops-MM embedding model; (ii) \textbf{Local FileVisit} for accessing relevant files and returning high-resolution images and structured PDF–Markdown content. 

We construct a stratified subset of 200 QAs (50\% reasoning, 30\% summary, 10\% induction, 10\% solution) guided by three principles: (1) \textbf{Objective evaluation} prioritizing verifiable logical structures; (2) \textbf{Multi-modal rigor} requiring integration of complex cross-document elements (e.g., figures); and (3) \textbf{Pipeline completeness} encompassing the full retrieval-to-synthesis workflow.
\subsubsection{Selected Baselines}
We evaluate baselines across two categories: (1) \textbf{MLLM-based ReAct agents}, including WebWatcher 32B~\cite{geng2025webwatcher}, kimi k2~\cite{team2025kimi}, GLM 4.5V~\cite{vteam2025glm45vglm41vthinkingversatilemultimodal}, Qwen3-VL~\cite{qwen3vl}, DeepSeek-V3.1~\cite{deepseekai2024deepseekv3technicalreport}, OpenAI 4o-mini~\cite{openai_4o_mini}, Gemini-2.5-flash-thinking, Gemini-2.5-pro~\cite{gemini_25_pro}, and OpenAI GPT-5.1; and (2) \textbf{Deep Research-specialized models}, such as DR Tulu-8B~\cite{shao2025dr}, MMSearch-R1-7B~\cite{wu2025mmsearch}, Asearcher-Web-7B~\cite{gao2025turnsunlockinglonghorizonagentic}, MiroThinker-v1.0-30B~\cite{miromind2025mirothinker}, Tongyi Deep Research 32B~\cite{tongyidr}, OpenAI o3 Deep Research~\cite{dr}, and Grok 4~\cite{grok4}. See Appendix~\ref{sec:appendix baseline} for details.

\begin{figure*}[h]
    \centering
    \includegraphics[width=1\textwidth]{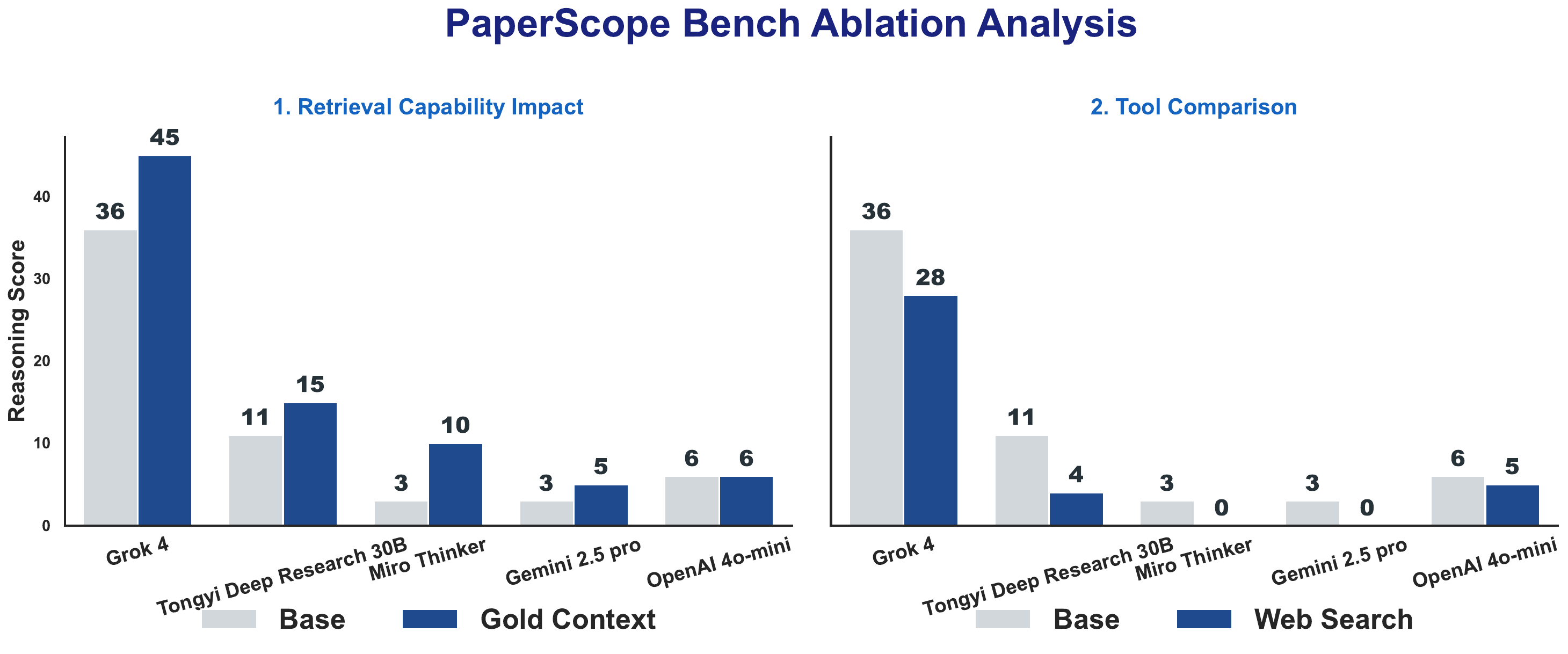}
    %\vspace{-1ex}
    \caption{The ablation results of \ours{} Bench in reasoning task.}
    %\vspace{-2ex}
    \label{fig:ablation study}
\end{figure*}

\subsection{Main Results}

For each task, macro-averaged scores are reported in Table~\ref{tab:agent_performance}, with more detailed results provided in Appendix~\ref{sec:appendix eval}. In addition, we present case studies of different model categories to illustrate systematic performance variations across task types, as shown in Figure~\ref{fig:modelcase}. Based on these results, several key observations can be drawn:
\begin{figure*}[h]
    \centering
     %\vspace{-2ex}
    \includegraphics[width=1\linewidth]{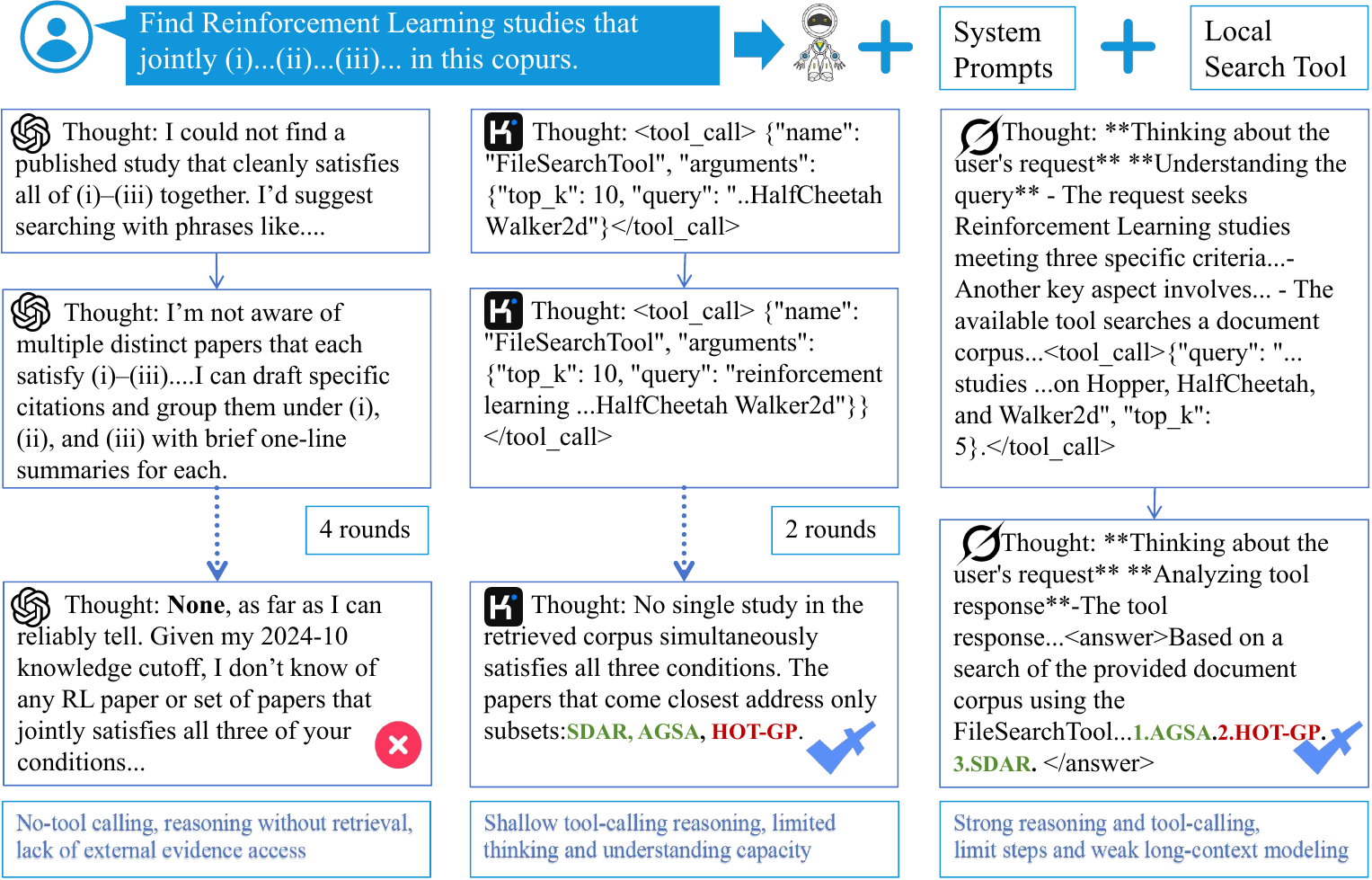}
    \caption{A case study comparing the capabilities of different model types in tool use and reasoning. The figure illustrates the behavioral distinctions and performance gaps between models with no tool support (left), shallow tool calling (middle), and strong reasoning combined with tool calling (right) when addressing complex retrieval tasks. For brevity, intermediate tool return results and the FileVisit tool interactions are omitted.}
    %\vspace{-2ex}
    \label{fig:modelcase}
\end{figure*}

(1) \textbf{Deep-research models show advantages in reasoning and tool-mediated integration.}
    In the heavily weighted Reasoning task, deep-research agents outperform LLM-based ReAct baselines, reflecting stronger capacity for cross-document and cross-modal evidence integration involving tables, figures, formulas, and algorithms.
    Representative models such as Grok-4 and OpenAI o3 deep research achieve comparatively high scores across Reasoning, Summarization, and Solution tasks. This advantage depends critically on stable tool-calling. However, several deep-research models and MLLMs fail to execute tools reliably, which substantially degrades their reasoning and induction performance.
    
(2) \textbf{ReAct models remain effective for summarization but fall short in scientific problem solving.}
    As shown in Table~3, ReAct-based systems such as Kimi~k2, Qwen3-VL, and DeepSeek-V3.1 achieve strong Summarization results.
    Their overall performance is nevertheless constrained by weak Reasoning and Solution scores, indicating limited ability to perform multi-source synthesis, cross-modal alignment, and structured inference required in scientific settings.

(3) \textbf{Complex multi-modal reasoning constitutes a universal bottleneck.} Most models struggle to interpret figures, tables, and mathematical expressions, leading to consistently low Reasoning scores. Observed failures include broken logical chains, misaligned evidence, and insufficient multi-modal fusion. The sensitivity of some deep-research models to tool usage further indicates overfitting in their tool-calling strategies, which partially explains the lower performance of Tongyi Deep Research 30B. These results highlight persistent challenges in visual--semantic reasoning, evidence chain construction, and systematic decomposition.

(4) \textbf{Long-context modeling limits performance across paradigms.} Across paradigms, Reasoning performance remains low, indicating difficulty in inferring research intent and managing heterogeneous inputs. Performance plateaus below 60 scores even in Summarization and Solution tasks, suggesting a reliance on parametric knowledge over reasoning. Extended multi-turn interactions frequently trigger catastrophic forgetting and hallucinations, which, combined with unreliable tool invocation, destabilize reasoning trajectories and constrain practical deployment.

\subsection{More Ablation Analysis}

(1) \textbf{Retrieval quality matters, but reasoning is the dominant bottleneck.}
        Directly providing oracle support documents improves performance (Figure~\ref{fig:ablation study} (1)), yet neither open-source nor closed-source models achieve strong results. This indicates that while retrieval is a factor, the primary bottleneck lies in the inherent complexity of the cross-source multi-modal reasoning required by the benchmark. More details are shown in Appendix~\ref{sec:appendix eval}.
    
(2) \textbf{Domain-specific local retrieval outperforms generic WebSearch.}
        To examine the impact of the provided local retrieval tool on model performance, we replace it with a commonly used community web search tool. As shown in Figure.\ref{fig:ablation study} (2), the evaluation scores of the tested models drop substantially. This decline can be attributed to the models’ inability to reliably retrieve the correct support documents when using web search, causing them to rely predominantly on knowledge encoded in their model parameters rather than evidence grounded in retrieved documents.

\begin{table*}[htbp]
\centering
\small
\begin{tabular}{@{} l c c p{4.5cm} @{}}
\toprule
\textbf{Setting} & \textbf{Combinations (Uniqueness)} & \textbf{Max Quality Score} & \textbf{Interpretation} \\
\midrule
\multicolumn{4}{c}{\textit{(a) Walk Length ($L$) --- finding the semantic sweet spot (with $W=10000, \beta=0.3$)}} \\
\midrule
$L=10$ & 2 & 60.33 & \textbf{Too Short:} Fails to escape local nodes. \\
\textbf{$L=100$ (Default)} & 10,000 (Max) & \textbf{1,564.19} & \textbf{Optimal:} Best balance of diversity and efficiency. \\
$L=500$ & 10,000 (Max) & 26,430.99 & \textbf{Diminishing Returns:} Deep links but introduces noise. \\
$L=1000$ & - & - & \textbf{Divergence:} Paths drift too far, failing to converge. \\
\midrule
\multicolumn{4}{c}{\textit{(b) Bias Probability ($\beta$) --- exploration vs. exploitation (with $L=100, W=10000$)}} \\
\midrule
\textbf{$\beta=0.3$ (Default)} & 10,000 (Max) & \textbf{1,564.19} & \textbf{Balanced:} Effectively explores long-tail entities. \\
$\beta=0.7$ & 10,000 (Max) & 1,789.55 & \textbf{Exploitation:} Favors hubs, slightly higher scores. \\
$\beta=1.0$ & 800 & 1,693.88 & \textbf{Collapse:} Fixates on hubs, destroying diversity. \\
\midrule
\textbf{Setting} & \textbf{Combinations (Uniqueness)} & \textbf{Time Cost} & \textbf{Interpretation} \\
\midrule
\multicolumn{4}{c}{\textit{(c) Number of Walks ($W$) --- coverage vs. cost (with $L=100, \beta=0.3$)}} \\
\midrule
$W=1000$ & 16 & $<$ 10s & \textbf{Under-sampling:} Sparse trajectories miss targets. \\
\textbf{$W=10,000$ (Default)} & 10,000 (Max) & \textbf{$\sim$ 60s} & \textbf{Efficiency Saturation:} Captures most targets quickly. \\
$W=50,000$ & 10,000 (Max) & $\sim$ 480s & \textbf{High Cost:} 8$\times$ slower for negligible diversity gain. \\
\bottomrule
\end{tabular}

\caption{Ablation analysis of ORWAS hyperparameters. Parameter tuning significantly impacts the uniqueness (discovery rate) of document combinations. Our default settings yield the maximum combination diversity without incurring excessive computational overhead.}
\label{tab:orwas_ablation}
\end{table*}
(3) \textbf{Optimal ORWAS hyperparameters balance combination diversity and computational cost.}
        We conducted a grid search on \texttt{corpus\_test} to evaluate the core hyperparameters of the ORWAS algorithm: Walk Length ($L$), Bias Probability ($\beta$), and Number of Walks ($W$). To measure effectiveness, we introduce a \textit{Combination Quality Score}---a weighted metric based on shared entity importance and coverage. A higher score implies tighter semantic coupling between papers, enabling human annotators to construct harder, more logical reasoning chains. 
        
        As shown in Table~\ref{tab:orwas_ablation}, our default settings ($L=100, \beta=0.3, W=10000$) maximize diversity (measured by unique combinations) at a low computational cost. Specifically, short walk lengths ($L=10$) fail to escape local nodes, while excessively long walks ($L \geq 500$) introduce noise or diverge (Table~\ref{tab:orwas_ablation} (a)). For bias probability, heavy exploitation ($\beta=1.0$) causes paths to collapse onto high-frequency hubs, drastically reducing uniqueness (Table~\ref{tab:orwas_ablation} (b)). Finally, $W=10,000$ reaches efficiency saturation in roughly 60 seconds; further scaling incurs higher time costs without significant diversity gains (Table~\ref{tab:orwas_ablation} (c)). It is important to note that while ORWAS determines the \textit{structural complexity} (i.e., the logic chain), the final \textit{solving difficulty} is primarily driven by the multi-modal information density (e.g., charts and tables) within the retrieved papers.

\subsection{Cross-Domain Generalizability}
To verify broader applicability, we conducted a pilot study on 24 papers from Medicine and Mechanics. By minimally adapting the node schema to domain-specific structures (e.g., adding \textit{Limitations} for Medicine and \textit{Equations} for Mechanics), our pipeline constructed high-density knowledge graphs ($>690$ nodes and $>35,000$ edges per domain) achieving a 95\% expert satisfaction rate. Furthermore, the framework successfully generated domain-specific multi-modal reasoning tasks, such as joint-chart analyses for clinical effect sizes and cross-verifying tabular data with theoretical mechanics formulas. Crucially, our automated random-walk algorithm seamlessly linked these cross-document entities and modalities without requiring domain-specific heuristic tuning. These results demonstrate our method's strong cross-domain transferability with adaptation costs.

\section{Conclusion}

In this paper, we present \ours{}, a comprehensive multi-modal, multi-document benchmark tailored for the challenges of scientific deep research. Constructed from knowledge graphs spanning over 2,000 AI papers, \ours{} synthesizes semantically aligned information nodes and employs random walk algorithms to sample seed papers. This process yields over 2,000 high-quality QA pairs encompassing 4 distinct capabilities: topic induction, reasoning, summarization, and solution generation. Experimental evaluations reveal that even state-of-the-art agents face significant challenges (Grok 4’s score of 40.95), underscoring the persistent difficulties in long-context retrieval and multi-document reasoning. Beyond serving as a evaluation suite and a scalable construction pipeline, we envision \ours{} as a pivotal resource for model training. 

\newpage
\section*{Acknowledgment}
This work was supported by National Natural Science Foundation of China No. 62272467. The work was partially done at the Beijing Key Laboratory of Research on Large Models and Intelligent Governance  and Engineering Research Center of Next-Generation Intelligent Search and Recommendation, MOE.
\section*{Limitations}

\ours{} is constructed on a large but finite local corpus rather than a fully open-domain setting, which may not fully reflect challenges arising from unbounded document collections and noisy retrieval.
Moreover, ORWAS adopts heuristic design choices, such as biased random walks and constrained combination enumeration, which improve efficiency but do not guarantee globally optimal article selection.
In addition, query and annotation generation rely on large language models with prompt-based control, where residual noise and model bias may affect evaluation stability.
Finally, due to computational constraints, experiments are conducted with limited baseline coverage and fixed inference budgets, leaving more extensive evaluations to future work.
% paperscope


\bibliography{custom}

\newpage

\appendix
\section{Detailed Overview of PaperScope}
\label{sec:appendix overview}
\subsection{Detailed Benchmark Statistic}

\ours{} consists of multiple large-scale corpus constructed from recent AI conference papers.
To facilitate reproducibility, we report detailed corpus-level statistics in Table~\ref{tab:corpus_stat}.

\begin{table}[h]
\centering
\small
\begin{tabular}{lcccc}
\hline
Corpus Split & \#Documents &  Source Venues & \#QA Pairs \\
\hline
Corpus-A & 500  & 5 conferences & 1100 \\
Corpus-B & 1500  & 5 conferences & 1100 \\
Test Corpus  & 202 & 5 conferences & 200 \\
\hline
\end{tabular}
\caption{High-level statistics of \ours{} corpus.}
\label{tab:corpus_stat}
\end{table}

\vspace{0.5em}
We further analyze the distribution of answer-support documents per question in Table~\ref{tab:answer_doc_dist},
which reflects the degree of multi-document dependency required by each task.

\begin{table}[h]
\centering
\small
\begin{tabular}{cccccc}
\hline
1 Doc & 2 Docs & 3 Docs & 4 Docs & 5+ Docs  \\
\hline
200 & 200 & 400 & 200 & 1400 &  \\
\hline
\end{tabular}
\caption{Distribution of the number of supporting documents per question.}
\label{tab:answer_doc_dist}
\end{table}

\subsection{Data Collection}
All documents utilized in this study are strictly sourced from repositories governed by the Creative Commons 4.0 (CC 4.0) licensing framework. Furthermore, the constructed queries were designed to focus exclusively on content analysis, ensuring the complete exclusion of personally identifiable information (PII) or sensitive private data regarding the authors.

\section{More Detailed of Data Construction}
\label{sec:appendix dataconstruction}
\subsection{Detailed Graph Statistic}

\begin{figure*}[h]
\centering
\includegraphics[width=1\textwidth]{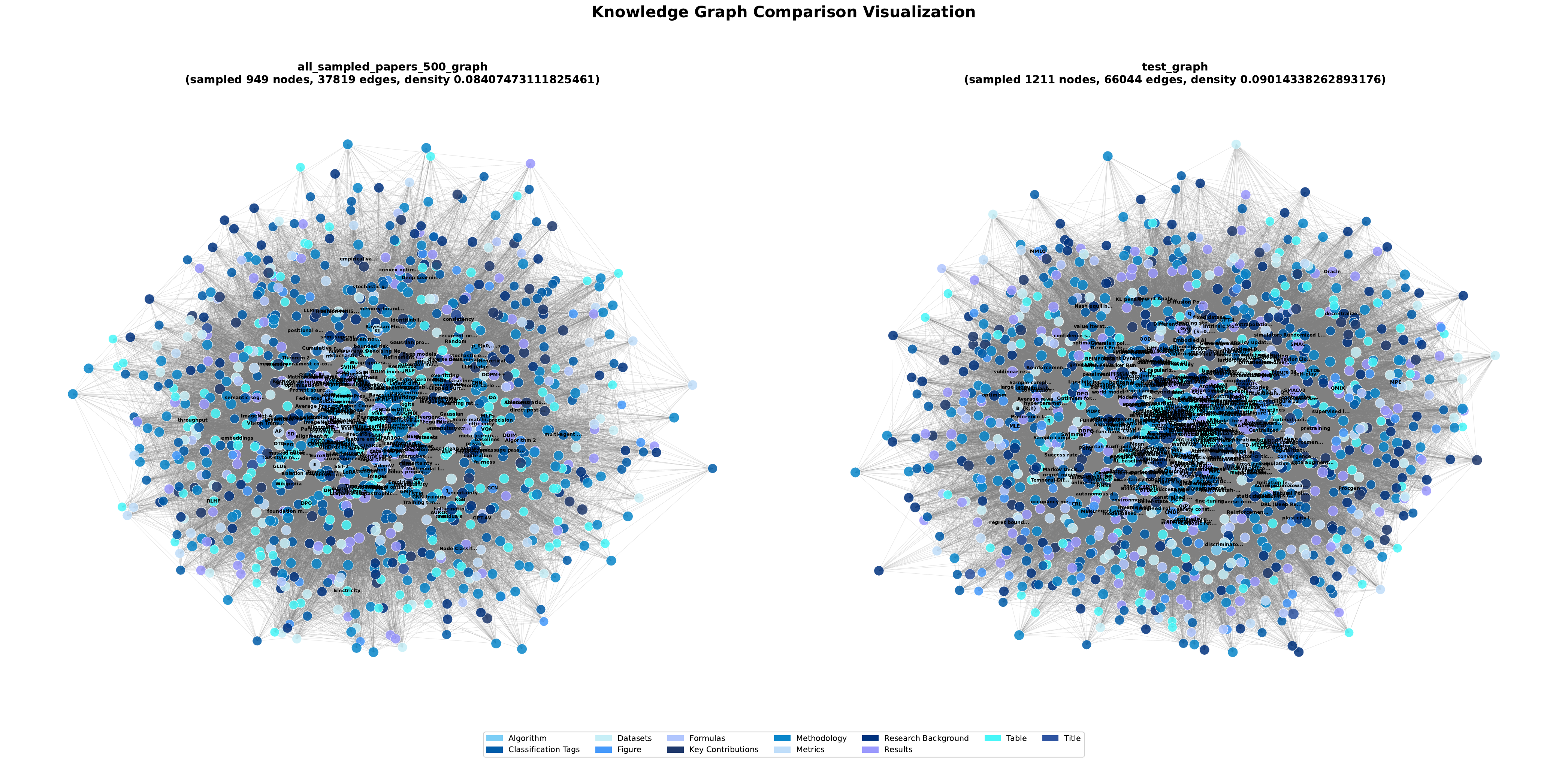}
\caption{Visualization of selected semantic graphs.}
\label{fig:node_type_dist}
\end{figure*}

We construct two large heterogeneous semantic graphs from different corpora.
Table~\ref{tab:graph_stat} summarizes their structural properties. Figure~\ref{fig:node_type_dist} visualizes the graphs.

\begin{table}[h]
\centering
\small
\resizebox{\columnwidth}{!}{
\begin{tabular}{lccc}
\hline
Statistic & Graph-Test & Graph-Docs 500  & Graph-Docs 1500\\
\hline
\#Nodes & 30962 & 31348 & 85152\\
\#Edges & 1990449 & 1907081 & 12735443\\
Density & 0.0042 & 0.0039 & 0.0025\\
Diameter & 3 & 4 & 8\\
Avg. Degree & 128.57 & 121.67 & 125.81\\
Avg. Path Length & 2.041 & 2.17 & 4.14\\
\hline
\end{tabular}
}
\caption{Structural statistics of the constructed semantic graphs.}
\label{tab:graph_stat}
\end{table}

\subsection{Detailed Optimized Random Walk Article Selector (ORWAS)}

ORWAS operates on a heterogeneous paper-key information node graph to identify compact yet thematically coherent paper subsets. Detailed core algorithm is shown in Algorithm~\ref{algr-stageA}. Table~\ref{tab:orwas_complexity} summarizes the time and space complexity.

\begin{table}[h]
\centering
\small
\begin{tabular}{lcc}
\hline
Stage & Time Complexity & Space Complexity \\
\hline
Random Walk & $O(W \cdot L)$ & $O(L)$ \\
Aggregation & $O(|A| \cdot |E|)$ & $O(|A|)$ \\
Combination Filter & $\min(C, \binom{N}{k})$ & Linear in output \\
\hline
\end{tabular}
\caption{Complexity analysis of ORWAS.}
\label{tab:orwas_complexity}
\end{table}

\begin{algorithm}[!t]
\caption{Stratified Random Walk Sampling}
\label{algr-stageA}
\textbf{Input:}
Heterogeneous graph $G=(V,E)$;
Article nodes $V_a$;
Key information nodes $V_e$;
High-frequency key information nodes $V_h$;
Walk length $L$;
Number of walks $W$;
Bias probability $\beta$ \\
\textbf{Output:}
Article--key information node map $\mathcal{M}$
\begin{algorithmic}[1]
    \State Initialize neighbor cache $\mathcal{N}(v)$ for all $v \in V$
    \State Initialize article--key information node map $\mathcal{M} \gets \emptyset$

    \State Sample $0.7W$ start nodes from $V_a$ and $0.3W$ from $V_h$
    \ForAll{start node $s$ \textbf{in parallel}}
        \State $path \gets [s]$
        \For{$i = 1$ to $L$}
            \State $u \gets$ last node in $path$
            \If{$u \in V_a$ \textbf{and} rand() $< \beta$}
                \State Select $v$ uniformly from $\mathcal{N}(u) \cap V_h$
                \If{empty}
                    \State Select $v$ uniformly from $\mathcal{N}(u)$
                \EndIf
            \Else
                \State Select $v$ uniformly from $\mathcal{N}(u)$
            \EndIf
            \State Append $v$ to $path$
        \EndFor
        \State Extract article--key information node pairs from $path$ and update $\mathcal{M}$
    \EndFor

    \State \textbf{return} $\mathcal{M}$
\end{algorithmic}
\end{algorithm}

% \begin{algorithm}[!t]
% \caption{Article Filtering by Shared Key Information}
% \label{algr-stageB}
% \textbf{Input:}
% Article--key information node map $\mathcal{M}$;
% Min shared key information nodes $\tau$ \\
% \textbf{Output:}
% Filtered article set $\mathcal{A}$
% \begin{algorithmic}[1]
%     \State $\mathcal{A} \gets \{ a \mid |\mathcal{M}[a]| \ge \tau \}$
%     \State Retain at most 500 articles in $\mathcal{A}$ to limit search space
%     \State \textbf{return} $\mathcal{A}$
% \end{algorithmic}
% \end{algorithm}

% \begin{algorithm}[!b]
% \caption{Constrained Article Combination Enumeration}
% \label{algr-stageC}
% \textbf{Input:}
% Filtered article set $\mathcal{A}$;
% Article--key information node map $\mathcal{M}$;
% Target article size $k$;
% Min shared key information nodes $\tau$;
% Max combinations $C_{\max}$ \\
% \textbf{Output:}
% Article combinations $\mathcal{R}$
% \begin{algorithmic}[1]
%     \State Initialize result set $\mathcal{R} \gets \emptyset$
%     \State Initialize counter $c \gets 0$

%     \ForAll{article combinations $\mathcal{C} \subset \mathcal{A}$ with $|\mathcal{C}| = k$}
%         \If{$c \ge C_{\max}$}
%             \State \textbf{break}
%         \EndIf
%         \State $E_{\cap} \gets \bigcap_{a \in \mathcal{C}} \mathcal{M}[a]$
%         \If{$|E_{\cap}| \ge \tau$}
%             \State Compute quality score for $\mathcal{C}$
%             \State Add $\mathcal{C}$ to $\mathcal{R}$
%         \EndIf
%         \State $c \gets c + 1$
%     \EndFor

%     \State \textbf{return} $\mathcal{R}$
% \end{algorithmic}
% \end{algorithm}

\subsection{Detailed Task Formulation}

\paragraph{Detailed Tasks' Prompt}

\begin{figure*}[h]
\centering
\includegraphics[width=0.9\textwidth]{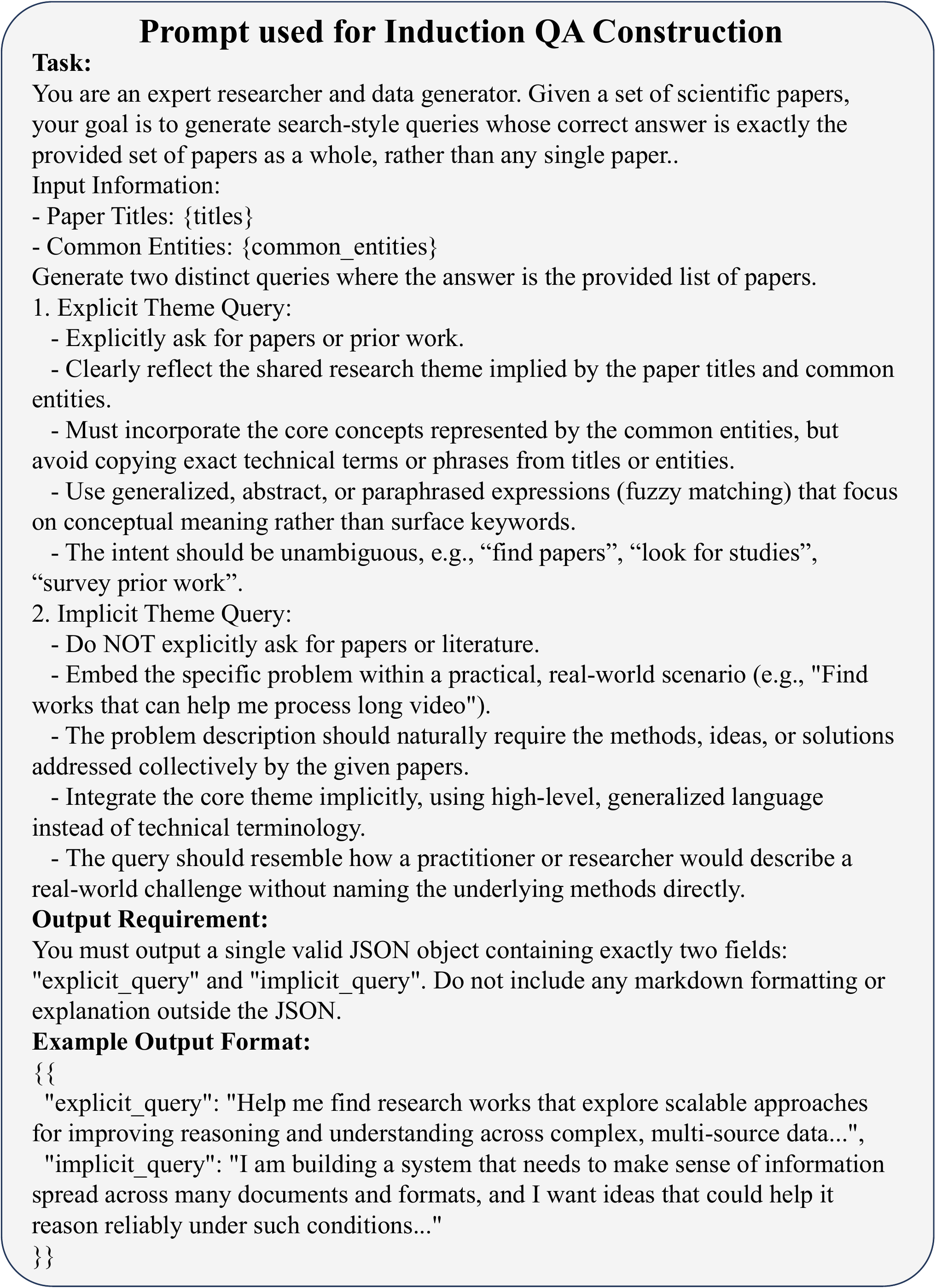}
\caption{The prompts used for induction task QAs construction.}
\label{fig:prompt1}
\end{figure*}

\begin{figure*}[h]
\centering
\includegraphics[width=0.9\textwidth]{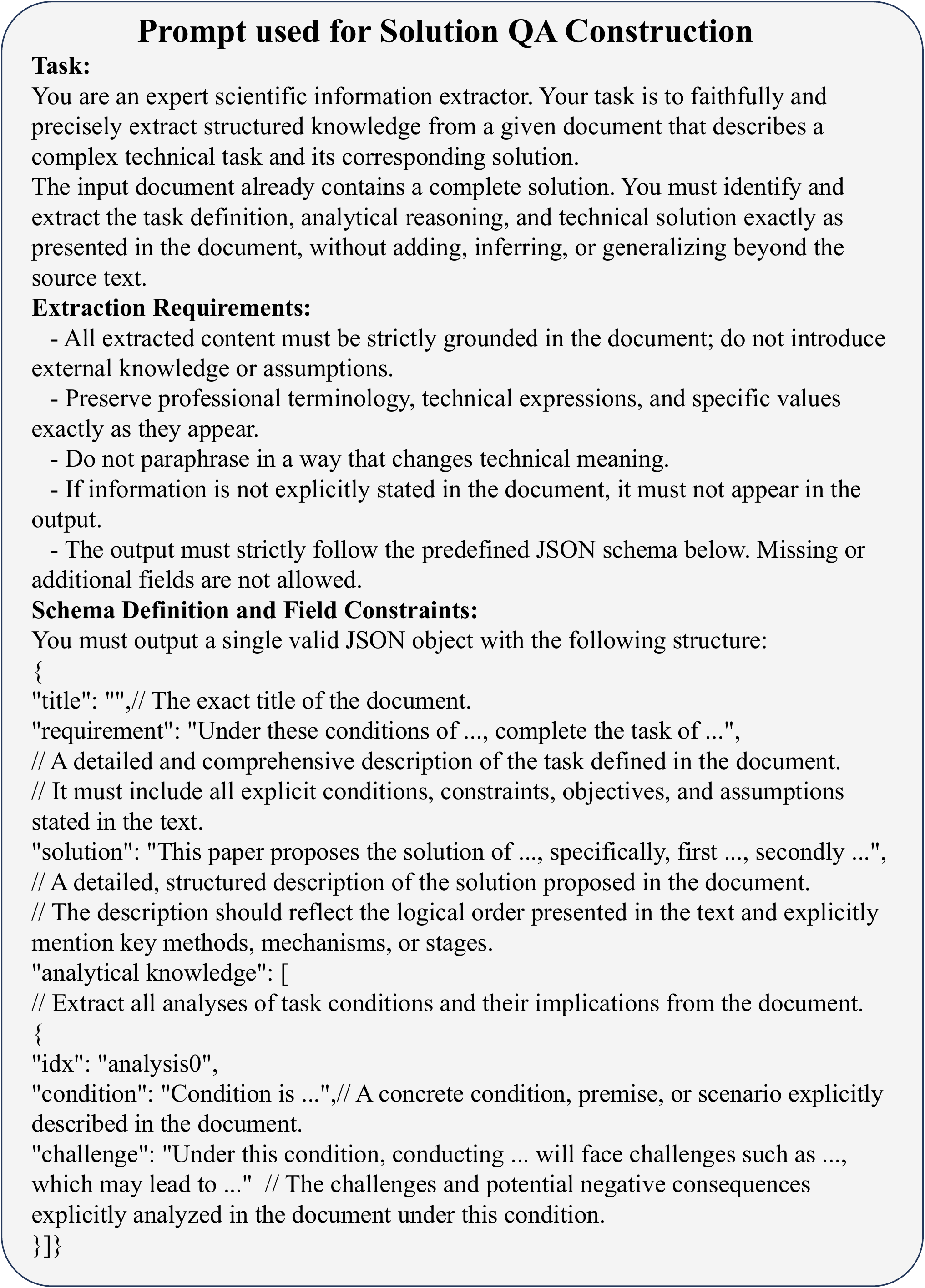}
\caption{The prompts used for solution task QAs construction.}
\label{fig:prompt2}
\end{figure*}

\begin{figure*}[h]
\centering
\includegraphics[width=0.9\textwidth]{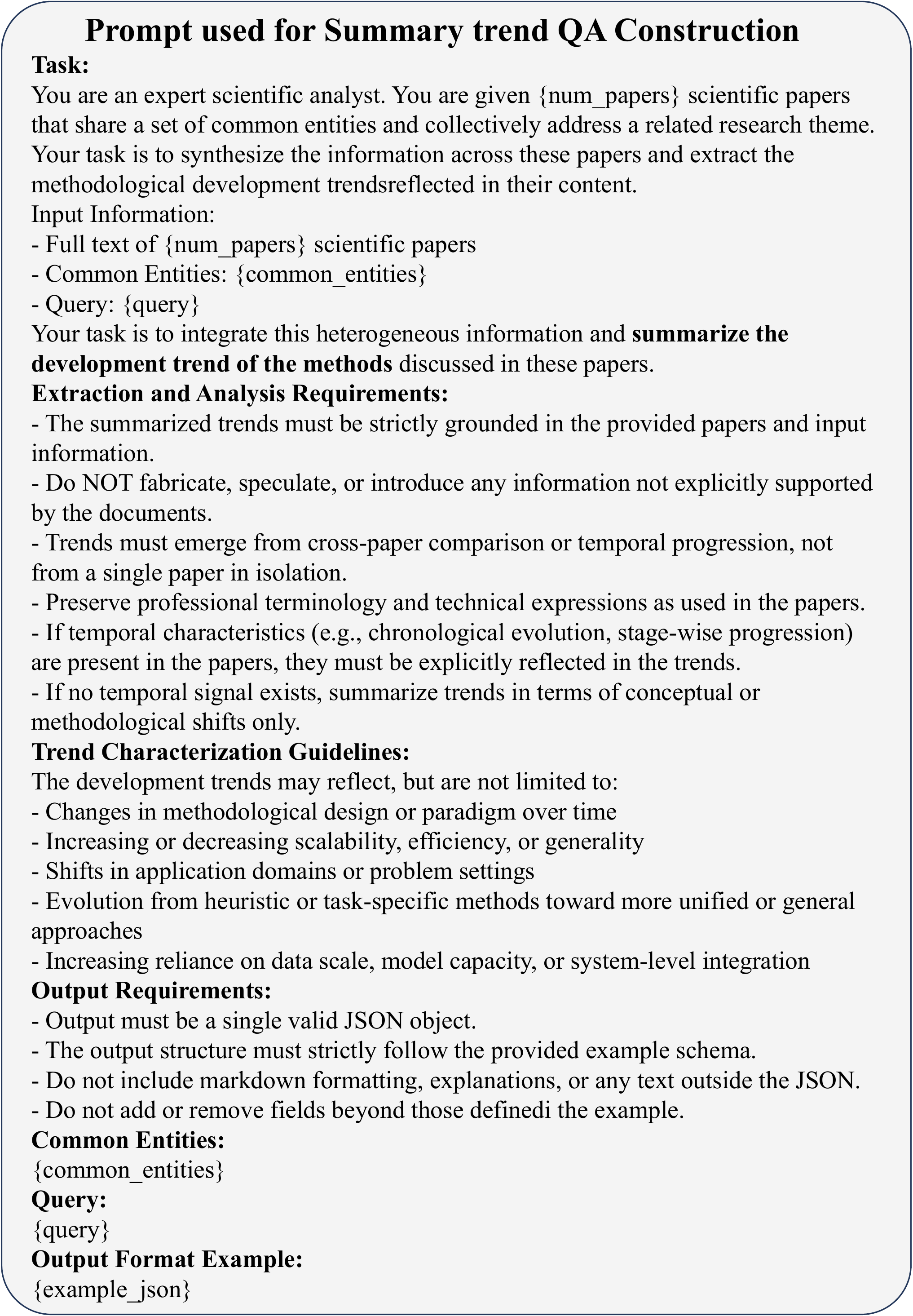}
\caption{The prompts used for summary trend task QAs construction.}
\label{fig:prompt3}
\end{figure*}

For the scientific reproducibility, we provide prompts for 3 types of synthetic tasks. Among them, the summary-type tasks include a dedicated prompt for trend questions, as shown in the Figure ~\ref{fig:prompt1},~\ref{fig:prompt2}, and~\ref{fig:prompt3}.

\subsection{Detailed Annotations}
To ensure the \ours{} benchmark maintains high intrinsic difficulty and strictly necessitates multi-modal reasoning, we implemented a bifurcated data construction strategy. This process distinguishes between expert annotation for complex reasoning tasks and a rigorous human-in-the-loop verification protocol for structured extraction tasks (Topic Induction, Summary, and Solution).
\subsubsection{Expert Annotation for Cross-Document Reasoning}
For the \textit{Reasoning} task category, computer science graduate students were engaged to formulate complex queries based on document clusters identified by the Optimized Random-Walk Article Selector (ORWAS). The annotation workflow was governed by an ``inverted construction'' paradigm designed to eliminate open-world ambiguity. Annotators first isolated irrefutable evidence nodes (ground truth) across multiple documents before deriving the corresponding questions, thereby guaranteeing \textit{Answer Uniqueness} and a closed retrieval scope.

A critical constraint imposed during this phase was \textit{Multi-modal Dependence}. Annotators were instructed to select evidence exclusively from visual modalities such as trend lines in ablation studies, numerical entries in comparative tables, or architectural connections in model diagrams. Questions were deemed valid only if they required cross-referencing these visual anchors with textual context; items solvable via text-only shortcuts were systematically rejected. Furthermore, to enforce \textit{Intrinsic Difficulty}, the protocol required that questions necessitate multi-hop inference. Queries resolvable through lexical matching or single-step web retrieval were discarded, ensuring the benchmark challenges the reasoning upper bounds of current state-of-the-art models.

\subsubsection{Expert Verification for Structured Tasks}
For \textit{Topic Induction}, \textit{Summary}, and \textit{Solution} tasks, we adopted an expert auditing mechanism to validate the quality of semi-automated candidates.
\begin{itemize}
    \item \textbf{Topic Induction Audit:} Reviewers verified that aggregated multi-modal nodes formed a coherent thematic query. The core criterion was exclusivity: the selected paper cluster must serve as the unique and optimal ground truth for the induced topic, ruling out weak or tangential associations.
    \item \textbf{Summary and Solution Audit:} For \textit{Summary} tasks, experts scrutinized the integration of fine-grained experimental comparisons with coarse-grained methodological trends, specifically checking for accurate temporal alignment and logical chart merging. In the \textit{Solution} category, validation focused on the structural integrity of the extracted knowledge, ensuring that specific conditions and challenges were logically organized under the unified theme without hallucination.
\end{itemize}

\subsection{Human-Model Evaluation Alignment}
\label{sec:human_model_alignment}
To demonstrate the reliability of our automated evaluation framework (particularly the GPT-5 judge), we conducted a blind Human-Model Alignment Study. We randomly sampled 10 instances each from the \textit{Summary} and \textit{Solution} tasks across 6 different models, yielding a total of 120 evaluation trajectories. Three CS Ph.D. students independently scored these trajectories using a streamlined rubric.

As detailed in Table~\ref{tab:alignment}, there is a strong and statistically significant positive correlation ($p < 0.0001$) between the automated evaluator and human judgments. According to established literature, a correlation coefficient surpassing 0.6 indicates substantial consensus. Notably, the automated judge performs exceptionally well on the reasoning-intensive \textit{Solution} task (Pearson $r = 0.6385$). Specifically, the correlation for the "Analysis" dimension peaked at 0.7173, indicating that the evaluator aligns closely with experts in identifying complex logical steps. Furthermore, its high Spearman correlation ($\rho > 0.6$) confirms strong ranking stability, validating its reliability as a diagnostic metric for agentic systems.

\begin{table}[htbp]
\centering
\resizebox{1\linewidth}{!}{
\begin{tabular}{@{}lccc@{}}
\toprule
\textbf{Task Type} & \textbf{Pearson ($r$)} & \textbf{Spearman ($\rho$)} & \textbf{P-value} \\
\midrule
Summary  & 0.6175 & 0.6072 & $< 0.0001$ \\
Solution & \textbf{0.6385} & \textbf{0.6263} & $< 0.0001$ \\
\bottomrule
\end{tabular}
}
\caption{Correlation between automated evaluation and human expert judgments based on 120 trajectories.}
\label{tab:alignment}
\end{table}

\subsection{Multi-modal Dependency and Data Verification}
\label{sec:multimodal_dependency}

\paragraph{Multi-modal Dependency.} 
To verify whether models genuinely engage in multi-modal reasoning rather than relying on textual shortcuts, we conducted a systematic Text-only Ablation experiment on a subset of 100 reasoning tasks. Visual inputs (e.g., charts and figures) were removed, leaving only OCR-extracted text. As shown in Table~\ref{tab:multimodal_ablation}, performance drops precipitously—by an average of 81.7\%—when visual inputs are absent. Only approximately 8\% of the questions could be solved via pure text (typically when specific values were explicitly described in the main text). This heavy visual dependency is driven by our "Visual Anchor" design strategy, which explicitly requires operations like cross-comparing a table's data with a figure's curve. Without visual perception, the logical chain inherently breaks.

\begin{table}[htbp]
\centering
\setlength{\tabcolsep}{4pt}
\resizebox{1\linewidth}{!}{
\begin{tabular}{@{}lccc@{}}
\toprule
\textbf{Model} & \textbf{Full Score} & \textbf{Text-only} & \textbf{Drop (\%)} \\
\midrule
WebWatcher 32B       & 4  & 1 & 75.0\% \\
Gemini 2.5 pro       & 3  & 1 & 66.7\% \\
Kimi k2              & 12 & 3 & 75.0\% \\
MiroThinker-v1.0-30B & 3  & 0 & \textbf{100.0\%} \\
OpenAI o3 (Deep Res) & 13 & 3 & 76.9\% \\
Grok 4               & 36 & 5 & \textbf{86.1\%} \\
\bottomrule
\end{tabular}
}
\caption{Text-only ablation results. The massive performance drop indicates that the benchmark heavily relies on visual information rather than simple text matching.}
\label{tab:multimodal_ablation}
\end{table}

\paragraph{Answer Uniqueness and Verification Statistics.}
In our benchmark, "uniqueness" refers to constraint-based determinability within the local corpus. To guarantee data quality, we implemented a rigorous dual-verification mechanism on Corpus-A:
\begin{itemize}
    \item \textbf{Stage 1: AI-based Difficulty Filtering.} We used an ensemble (Gemini-2.5-pro + GPT-5 + Web Search) to filter out trivial instances solvable via direct retrieval. This removed 9.98\% of the samples (112 items).
    \item \textbf{Stage 2: Human Validity Verification.} Expert annotators performed a secondary review to check for exclusivity, semantic coherence, and structural completeness. This filtered an additional 7.45\% (82 items), ensuring high reliability and unambiguous reasoning bounds for the retained tasks.
\end{itemize}

\section{More Detailed of baseline}
\label{sec:appendix baseline}

All experiments are conducted on a multi-GPU environment.
Table~\ref{tab:baseline_config} summarizes the default configuration.

\begin{table}[h]
\centering
\small
\begin{tabular}{lc}
\hline
Component & Configuration \\
\hline
GPUs & 8 $\times$ H100 \\
Inference Engine & vLLM \\
Max retries & 10 \\
Max excute time & 2h30mins \\
Max tokens & 128000 \\
Top p & 0.95 \\
Temperature & 0.6 \\
Batch Size & 4 \\
presence penalty & 1.1 \\
\hline
\end{tabular}
\caption{Experimental configuration for baseline evaluation.}
\label{tab:baseline_config}
\end{table}

\section{Detailed Evaluation of PaperScope}
\label{sec:appendix eval}
\subsection{Evaluation Prompts}

\begin{figure*}[h]
\centering
\includegraphics[width=0.9\textwidth]{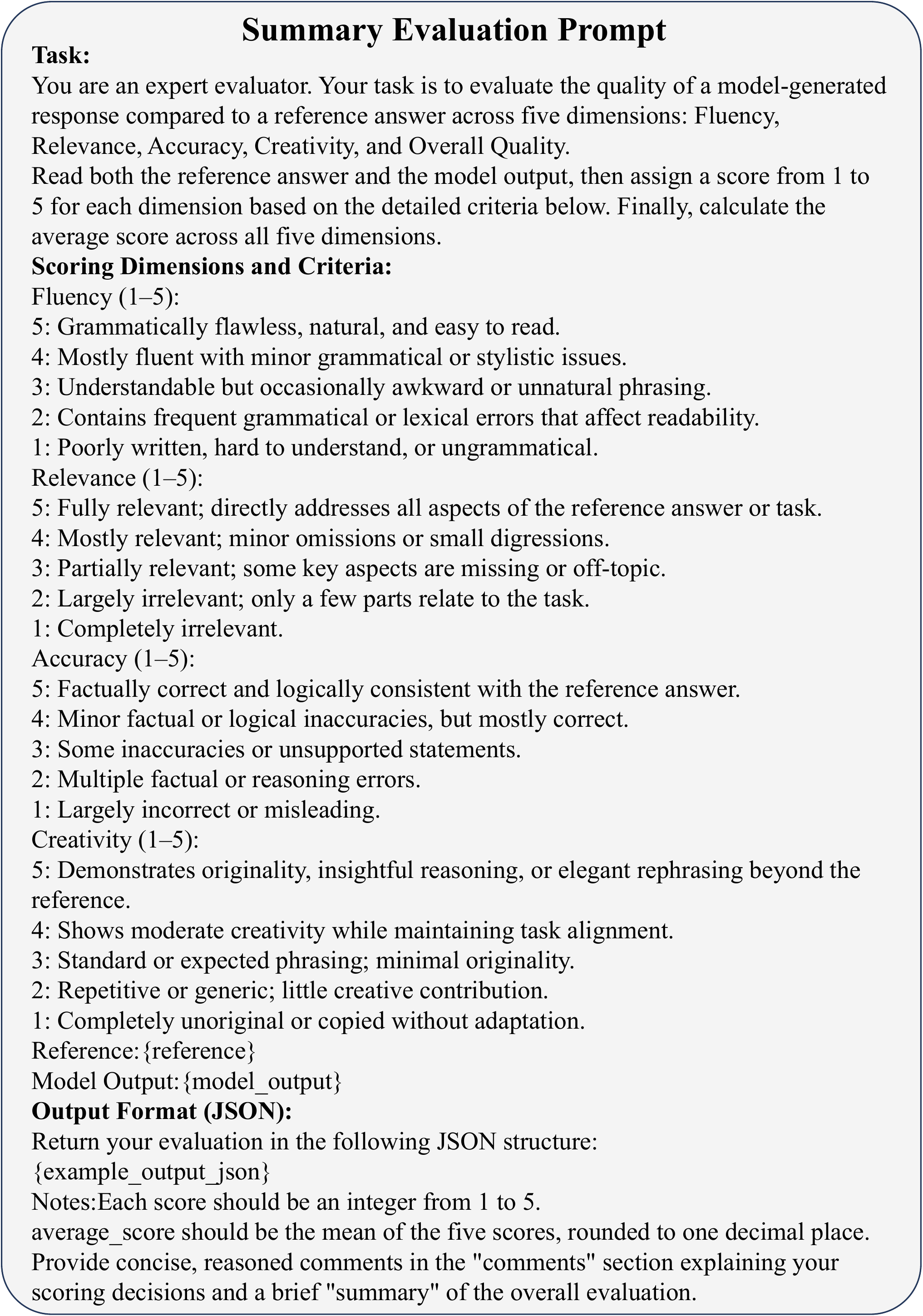}
\caption{The prompts used for summary task evaluation.}
\label{fig:prompts sum eval}
\end{figure*}

\begin{figure*}[h]
\centering
\includegraphics[width=0.9\textwidth]{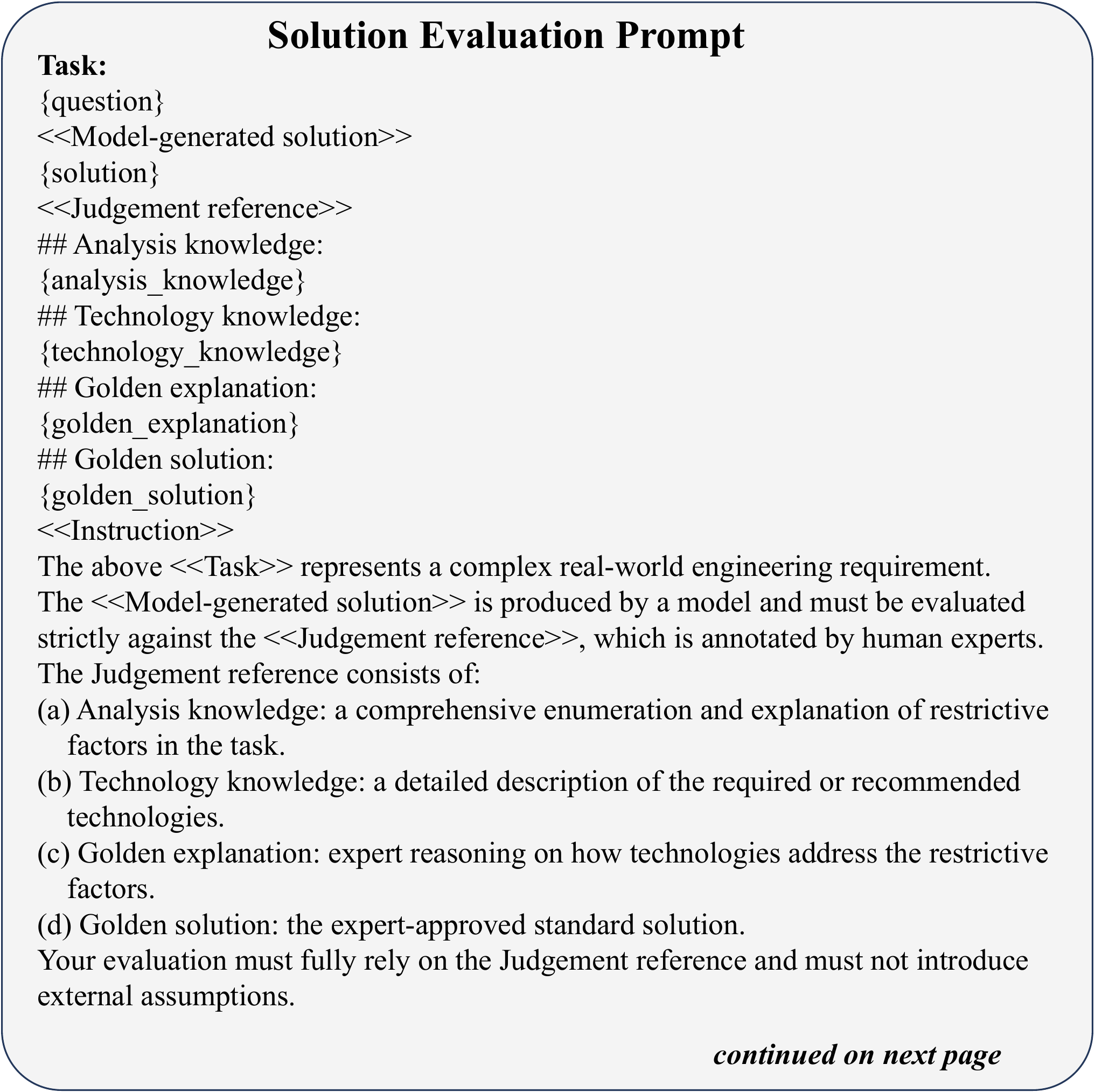}
\caption{The prompts used for solution task evaluation.}
\label{fig:prompts solu eval p1}
\end{figure*}

\begin{figure*}[h]
\centering
\includegraphics[width=0.9\textwidth]{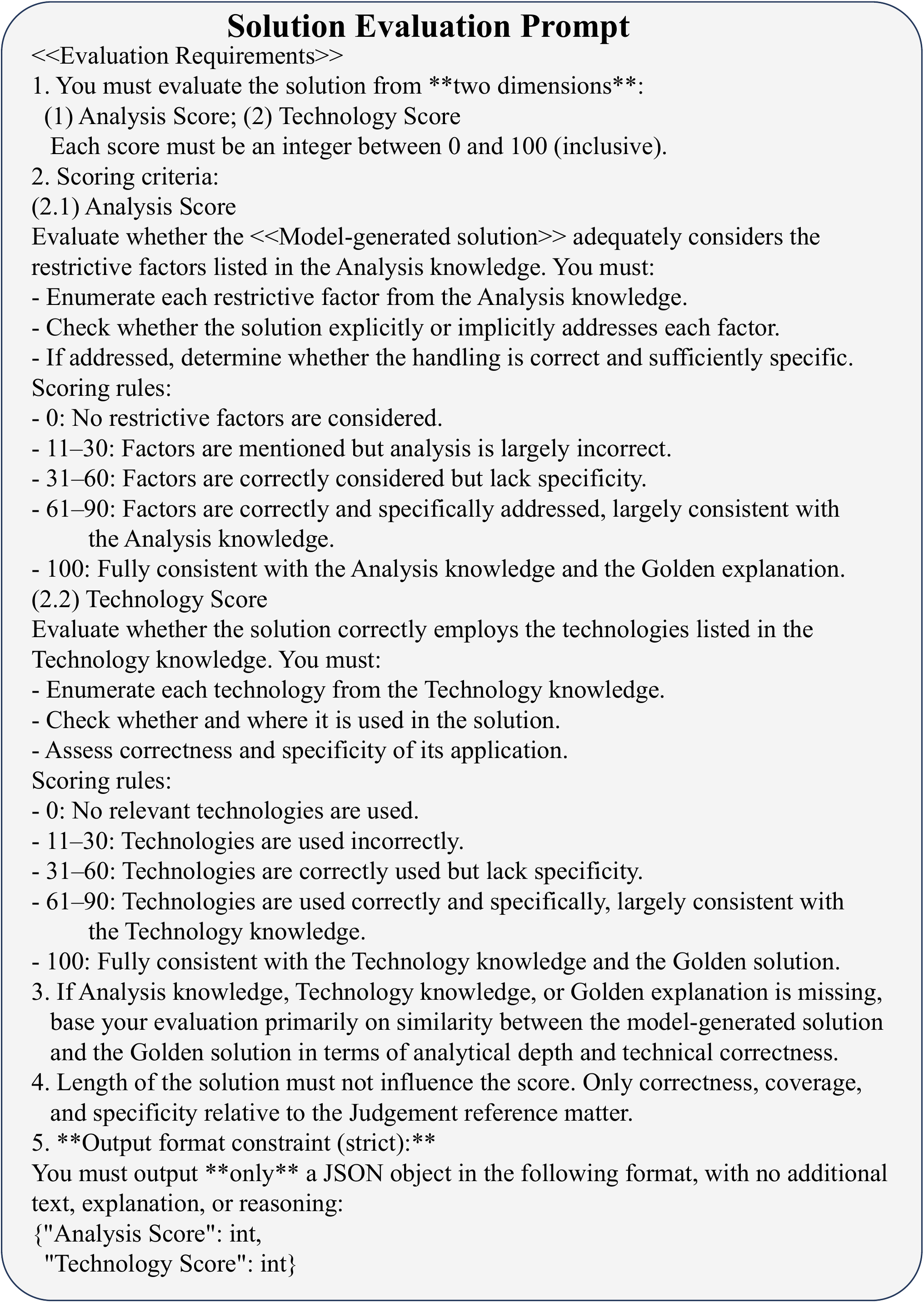}
\caption{The prompts used for solution task evaluation.}
\label{fig:prompts solu eval p2}
\end{figure*}

The specific evaluation prompts utilized for the Summarization and Solution tasks are delineated in Figure~\ref{fig:prompts sum eval}, Figure~\ref{fig:prompts solu eval p1} and ~\ref{fig:prompts solu eval p2}. 

For the \textbf{Summary task}, performance is assessed across five distinct dimensions: \textit{Fluency}, \textit{Relevance}, \textit{Accuracy}, \textit{Creativity}, and \textit{Overall Quality}. In this process, GPT-5 serves as the adjudicator, evaluating the efficacy of system-generated responses against gold-standard references based on these metrics.

Regarding the \textbf{Solution task}, evaluation is quantified via two primary metrics: the \textit{Analysis Score} and the \textit{Technology Score}. As illustrated in the Figure~\ref{fig:prompts solu eval p1} and ~\ref{fig:prompts solu eval p2}, we synthesized tailored evaluation rubrics for each component. To ensure robust assessment, the evaluator is supplied with a comprehensive context window comprising relevant analysis knowledge, technical knowledge, golden explanation and golden solution extracted from the source text.

\subsection{Detailed Tools}
The proposed \texttt{FileSearchTool} module is designed to construct an efficient document retrieval system based on semantic similarity, capable of processing a heterogeneous corpus containing Markdown text and various image formats (e.g., PNG, JPG). The core logic is encapsulated within the \texttt{FileSearchEngine} class, which orchestrates multi-modal embedding and Optical Character Recognition (OCR) technologies to achieve unified indexing and retrieval across modalities.

In terms of model architecture, the system integrates two pivotal pre-trained models. First, the \texttt{Ops-MM-embedding-v1}\footnote{Ops-MM-embedding-v1-7B: https://huggingface.co/OpenSearch-AI/Ops-MM-embedding-v1-7B} model is employed as a unified feature encoder, ensuring that text and image inputs possess commensurability within the same high-dimensional space. Second, the \texttt{DeepSeek-OCR}~\cite{wei2025deepseek} model is introduced for deep semantic understanding and text extraction. To optimize efficiency, the system utilizes the FAISS library to construct an \texttt{IndexFlatIP} index based on inner product similarity. During index construction, text content is extracted directly, while visual information is transformed into vector representations and stored alongside text embeddings to support unified retrieval.

In the retrieval phase, user queries are encoded to identify the top-$k$ nearest candidates. The system applies modality-aware post-processing: text documents are returned directly, while images undergo OCR and captioning via the prompt ``\texttt{<image>\textbackslash nCaption this image.}'' to generate textual descriptions. Integrated into the Qwen-Agent framework as a \texttt{BaseTool}, the module processes JSON parameters and returns aggregated content to serve as context augmentation for downstream Large Language Model tasks.

\subsection{Detailed Ablation Experiments}

To rigorously decouple the contribution of the retrieval module from the reasoning capabilities of the agent, and to assess the benchmark's resilience against generic information retrieval methods, we conducted a series of ablation experiments. These experiments were exclusively performed on the \textit{Reasoning} task subset. This specific subset was selected as the experimental substrate due to its strict multi-modal dependencies and high intrinsic difficulty, combined with the objectivity of its ground truth, which allows for unambiguous performance evaluation.
\subsubsection{Impact of Retrieval Necessity: The Oracle Setting}
To investigate whether the agent's performance is bottlenecked by retrieval accuracy or reasoning limitations, we implemented an experimental setup that bypasses the active search process. In this configuration, the standard retrieval module was replaced by a deterministic mechanism, referred to as the \textit{Direct Context Tool}. This tool was exposed to the model with the following functional description:
\begin{quote}
    \textit{``Directly retrieve pre-defined documents based on the original question without any retrieval process. Returns the parsed markdown documents corresponding to the question.''}
\end{quote}
By feeding the model the ground-truth parsed markdown documents directly, this setup simulates an ``oracle'' retrieval scenario. Comparing the performance of this configuration against the full pipeline allows us to quantify the gap between the model's reasoning potential given ideal context and its actual performance when burdened with the noise and uncertainty of retrieval.
\subsubsection{Impact of Generic Web Search}
To validate the hypothesis that the \ours{} benchmark requires specialized, domain-specific indexing rather than general open-web knowledge, we evaluated an agent configuration equipped with a commercial-grade search engine. We integrated the Bocha API to construct a \textit{Web Search Tool}, which was defined for the agent as follows:
\begin{quote}
    \textit{``Performs batched web searches: supply an array `query'; the tool retrieves the top 10 results for each query in one call.''}
\end{quote}
This ablation assesses the vulnerability of the benchmark to existing commercial search solutions. Under the assumption that the ``Answer Uniqueness'' and ``Multi-modal Dependence'' principles (Section~\ref{quality control}) were effectively implemented, the generic web search is expected to underperform, as the specific visual evidence and cross-document inference paths required for the \textit{Reasoning} tasks are unlikely to be indexed or synthesized effectively by standard search engines.

\subsection{Detail Evaluation Results}
To provide a more fine-grained characterization of model behavior across different task types, we report detailed results for the induction and solution tasks in which the models achieve non-zero performance, as shown in Table~\ref{tab:solution_model_scores} and Table~\ref{tab:induction_model_scores}. These task-specific analyses are intended to expose systematic performance patterns that are otherwise obscured by aggregate metrics. Due to space constraints, only partial results covering all subtasks are presented for a subset of models, as shown in Table~\ref{tab:some_model_detailed_performance}.

\begin{table*}[t]
    \centering
    \small
{%
        \begin{tabular}{l c c c}
            \toprule
            \textbf{Model} & \textbf{Max Score} & \textbf{Min Score} & \textbf{Macro Avg.} \\
            \midrule
            OpenAI o3 deep research & 79 & 35 & 59.15 \\
            kimi-k2 & 73 & 11 & 49.85 \\
            Grok-4 & 65 & 32 & 48.27 \\
            Gemini-2.5-pro & 51.5 & 19 & 40.39 \\
            DR Tulu-8B & 51 & 0 & 38.71 \\
            Qwen3-VL  & 59 & 25 & 37.38 \\
            Tongyi-DeepResearch-30B & 65 & 14 & 36.55 \\
            GLM 4.5V & 52 & 9 & 32.45 \\
            MiroThinker-v1.0-30B & 100 & 0 & 32.02 \\
            Gemini-2.5-flash-thinking & 62 & 0 & 29.70 \\
            WebWatcher-32B & 48 & 0 & 26.77 \\
            OpenAI 4o-mini & 42.5 & 9 & 22.10 \\
            MMSearch-R1-7B & 98 & 0 & 17.21 \\
            deepseek-V3.1 & 64.5 & 0 & 14.18 \\
            OpenAI GPT-5.1 & 61.5 & 0 & 12.95 \\
            ASearcher-Web-7B & 93.5 & 0 & 8.95 \\
            \bottomrule
        \end{tabular}%
    }
    \caption{Performance comparison of models on solution tasks showing Max Score, Min Score, and Macro Average.}
    \label{tab:solution_model_scores}
\end{table*}

\begin{table*}[t]
    \centering
    \small
    % 保持与上一表格相同的列间距设置，防止字体过大
    {%
        \begin{tabular}{l c c c}
            \toprule
            \textbf{Model} & \textbf{Explicit Topic} & \textbf{Implicit Topic} & \textbf{Macro Avg.} \\
            \midrule
            deepseek-V3.1 & 26.67 & 25.93 & 26.32 \\
            OpenAI 4o-mini & 18.52 & 33.33 & 25.49 \\
            kimi-k2 & 20 & 29.17 & 24.07 \\
            Grok-4 & 20 & 20 & 20 \\
            Gemini-2.5-flash-thinking & 13.33 & 13.33 & 13.33 \\
            Qwen3-VL & 16.67 & 10 & 13.33 \\
            Gemini-2.5-pro & 3.33 & 11.11 & 7.02 \\
            MiroThinker-v1.0-30B & 3.7 & 4.17 & 3.92 \\
            MMsearch-r1-7b & 3.7 & 3.7 & 3.7 \\
            \bottomrule
        \end{tabular}%
    }
    \caption{Performance comparison of models on Explicit Topic and Implicit Topic tasks.}
    \label{tab:induction_model_scores}
\end{table*}
\begin{table*}[t]
    \centering
    \small %稍微减小字号
    \renewcommand{\arraystretch}{1.2}
    \begin{tabular}{l ccccc}
        \toprule
        \textbf{Task} & \textbf{Gemini-2.5} & \textbf{GPT-5.1} & \textbf{MM-r1} & \textbf{Tulu-8b} & \textbf{Miro-30B} \\
        \midrule
        \multicolumn{6}{l}{\textit{\textbf{Reasoning}}} \\
        \quad Algorithm       & 5    & 0 & 10 & 6  & 5 \\
        \quad Fig-Tab-Cha Com & 0    & 0 & 0  & 10 & 0 \\
        \quad Fig-Tab-Cha Rea & 10   & 0 & 5  & 5  & 5 \\
        \quad Formula         & 10   & 0 & 20 & 0  & 5 \\
        \quad Paper           & 10   & 0 & 5  & 0  & 0 \\
        \midrule
        \multicolumn{6}{l}{\textit{\textbf{Induction}}} \\
        \quad Explicit        & 13.3 & 0 & 3.7 & 0 & 3.7 \\
        \quad Implicit        & 13.3 & 0 & 3.7 & 0 & 4.2 \\
        \midrule
        \multicolumn{6}{l}{\textit{\textbf{Summary}}} \\
        \quad Trend           & 41.0 & 50.6 & 44.2 & 51.8 & 29.8 \\
        \quad Dev.            & 44.0 & 50.6 & 51.6 & 49.8 & 31.8 \\
        \quad Comparison      & 52.2 & 55.2 & 49.8 & 49.8 & 28.8 \\
        \midrule
        \textbf{Solution}     & \textbf{29.7} & \textbf{12.95} & \textbf{17.21} & \textbf{38.71} & \textbf{32.02} \\
        \bottomrule
    \end{tabular}
    \caption{Performance comparison (Transposed View). Note: Model names are abbreviated for brevity.}
    \label{tab:some_model_detailed_performance}
\end{table*}

\subsection{Case Study}

We present one successful case and one failure case to illustrate typical agent behaviors.

\paragraph{Successful Case.}
\begin{figure*}[h]
\centering
\includegraphics[width=0.9\textwidth]{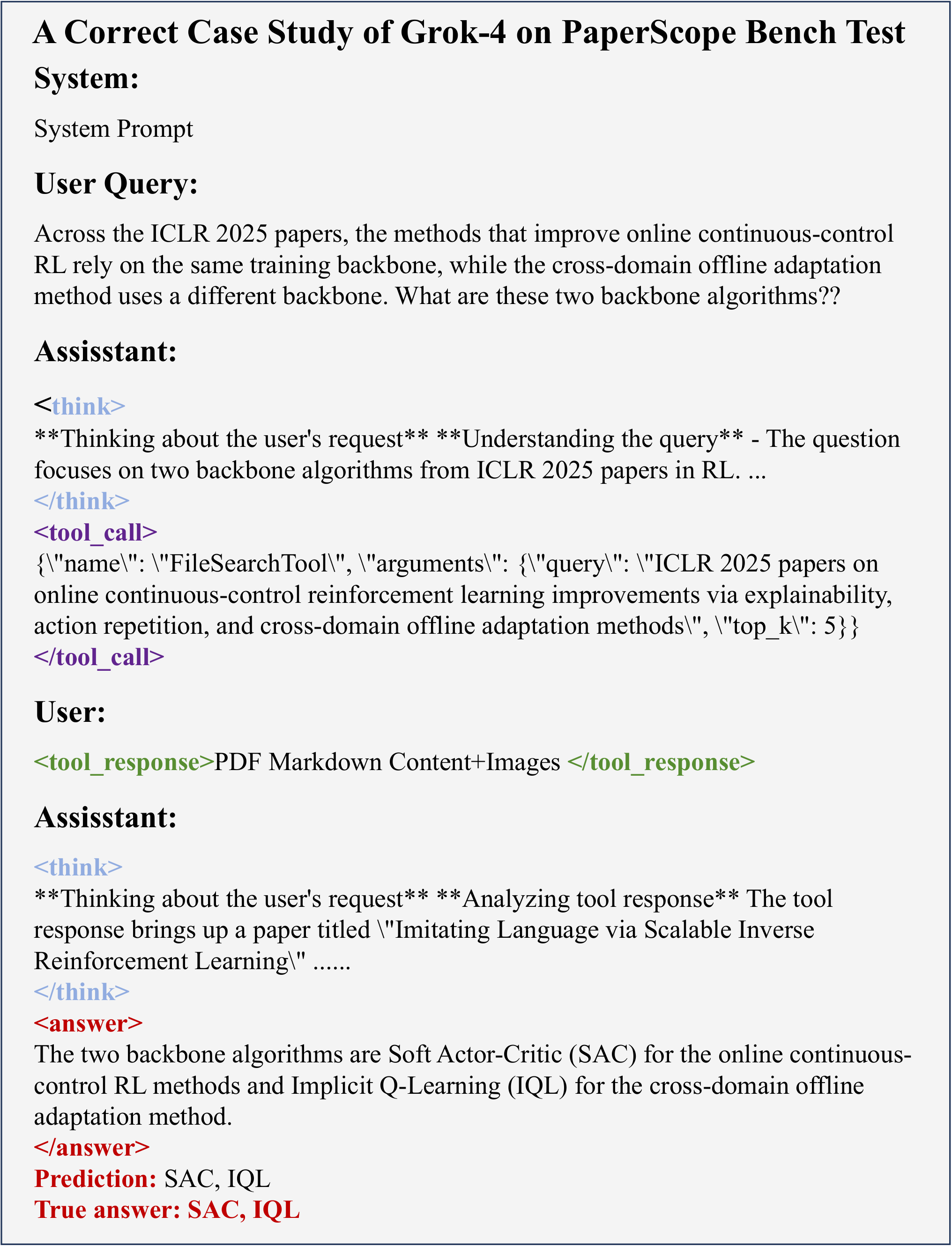}
\caption{a case study of Grok-4.}
\label{fig:correct case 1}
\end{figure*}
Figure~\ref{fig:correct case 1} presents a correct case which grok-4 successfully answer the full paper reasoning question.

\paragraph{Failure Case.}

\begin{figure*}[h]
\centering
\includegraphics[width=0.9\textwidth]{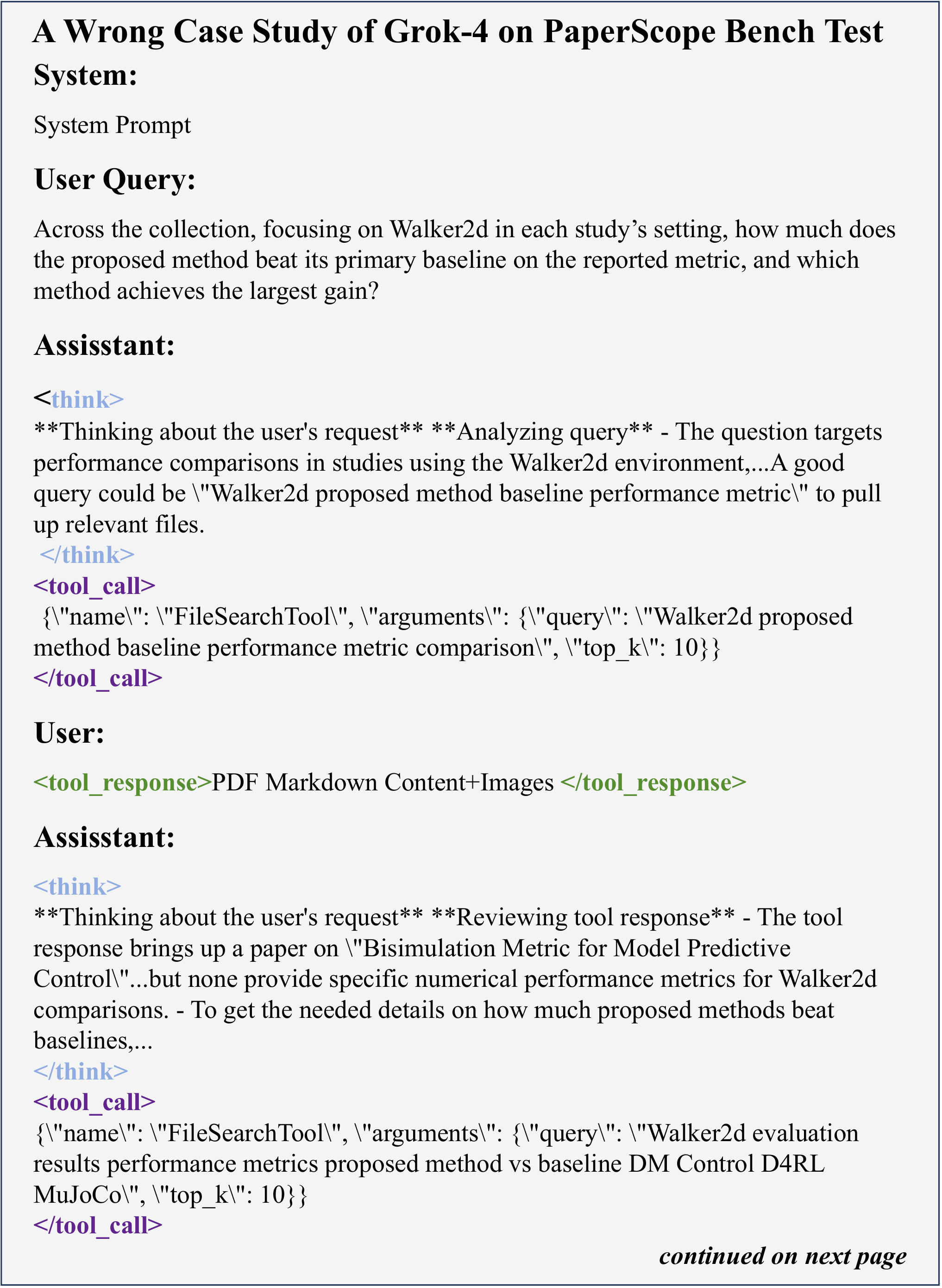}
\caption{a case study of Grok-4.}
\label{fig:wrong case p1}
\end{figure*}

\begin{figure*}[h]
\centering
\includegraphics[width=0.9\textwidth]{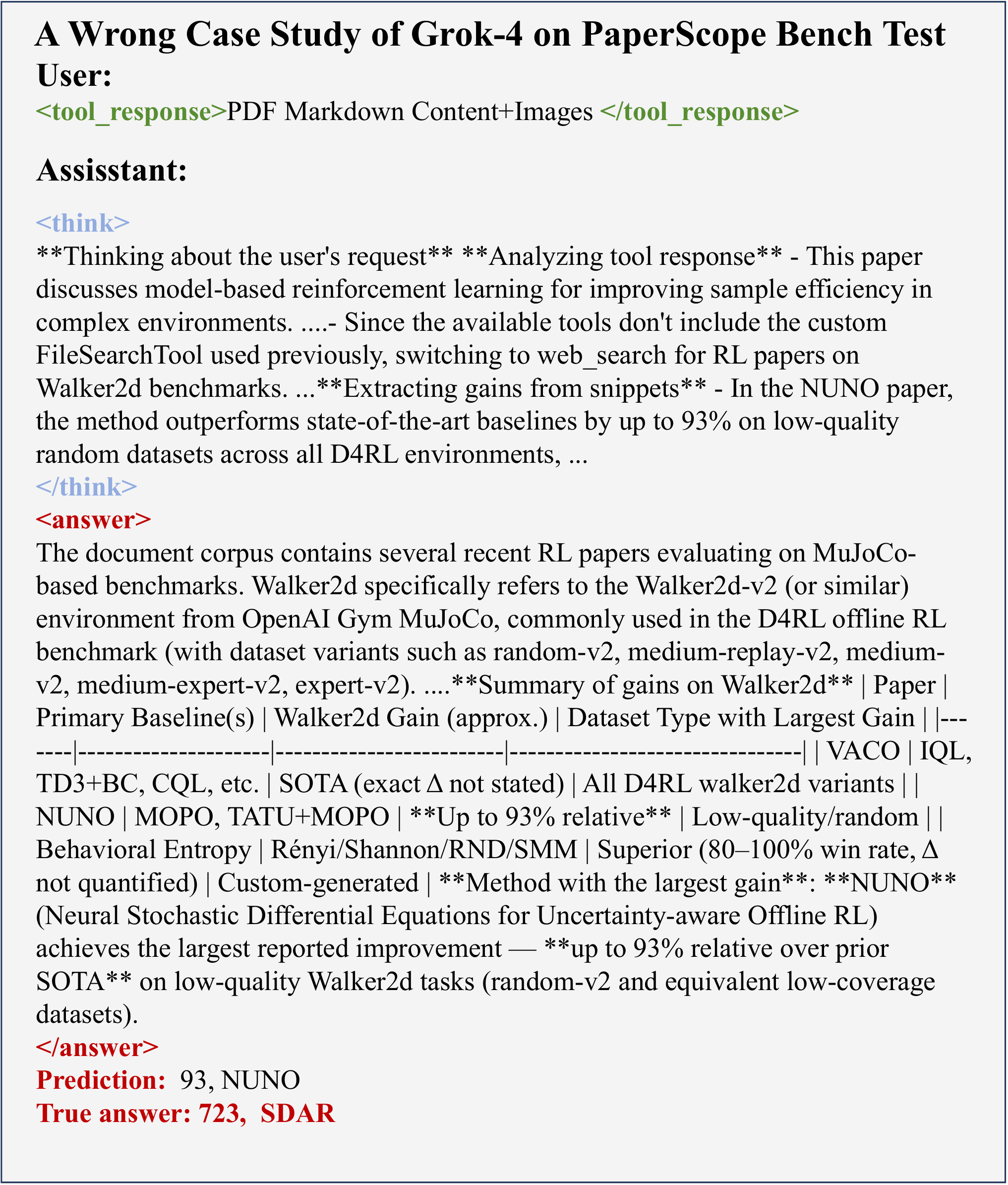}
\caption{a case study of Grok-4.}
\label{fig:wrong case p2}
\end{figure*}
Figure~\ref{fig:wrong case p1} and Figure~\ref{fig:wrong case p2} show a wrong case which grok-4 didn't retrieval the all golden documents and understanding all results in the tables and charts.

\section{Error Attribution Analysis}
\label{sec:error_analysis}
To understand exactly where agentic pipelines fail on complex academic tasks, we conducted a fine-grained Error Attribution Analysis on the strongest baseline model (Grok-4). Human experts manually reviewed 40 randomly sampled failure cases against the gold-standard documents. The breakdown of failure modes is presented in Table~\ref{tab:error_analysis}. 

\begin{table*}[htbp]
\centering
\small
\resizebox{\linewidth}{!}{
\begin{tabular}{@{}lp{5.5cm}cr@{}}
\toprule
\textbf{Failure Mode} & \textbf{Definition} & \textbf{Count} & \textbf{Ratio} \\
\midrule
Extraction Error & Correct document retrieved, but unable to read precise numerical values from charts/tables. & 15 & \textbf{37.5\%} \\
Reasoning Error  & Values extracted correctly, but multi-step calculation or logical inference fails. & 9  & 22.5\% \\
Retrieval Failure& Fails to search (7.5\%) or retrieves the wrong document (12.5\%). & 8  & 20.0\% \\
Hallucination    & Outputs speculative guesses despite successful retrieval. & 6  & 15.0\% \\
Context Limit    & Key information truncated due to long multi-document concatenation. & 2  & 5.0\%  \\
\bottomrule
\end{tabular}
}
\caption{Error attribution for Grok-4 based on 40 failure cases.}
\label{tab:error_analysis}
\end{table*}

Based on this analysis, we identify three major bottlenecks in current systems:
\begin{enumerate}
    \item \textbf{multi-modal Information Extraction is the Primary Weakness (37.5\%):} Surprisingly, fine-grained visual understanding bottlenecks the pipeline earlier than reasoning. Even when the correct paper is retrieved, the model often "reads but does not understand" (e.g., failing to align rows/columns in complex tables or extracting inaccurate coordinates from line charts).
    \item \textbf{Error Accumulation in Multi-step Reasoning (22.5\%):} In cross-paper synthesis tasks, models frequently err in intermediate steps. Small deviations in initial metric extraction compound during comparisons, causing final answers to drift significantly.
    \item \textbf{Retrieval Granularity and Hallucination (35\% combined):} Approximately 20\% of errors stem from broad semantic search scopes failing to pinpoint specific papers. The 15\% hallucination rate is often a secondary effect—when exact evidence is missed, models tend to generate speculative answers rather than abstaining.
\end{enumerate}

\section{Usage of LLM}
In the preparation of this manuscript, Large Language Models (LLMs) were utilized to facilitate translation and linguistic refinement. regarding the computational implementation, while LLMs assisted with auxiliary coding routines, the development of critical data processing pipelines and core algorithmic architectures was conducted exclusively by the authors to ensure scientific integrity.

\end{document}